%% file: main.tex
\documentclass[10pt,journal,compsoc,twocolumn]{IEEEtran}

\ifCLASSOPTIONcompsoc
  \usepackage[nocompress]{cite}
\else
  \usepackage{cite}
\fi
\ifCLASSINFOpdf
  \usepackage[pdftex]{graphicx}
\else
\fi
\usepackage{amsmath}
\usepackage{amssymb}
\usepackage{booktabs}
\usepackage{multirow}

\usepackage{enumitem}
\usepackage{multicol}
\usepackage[hidelinks]{hyperref}
\usepackage{cleveref}
\hypersetup{
  colorlinks   = true,
  urlcolor     = blue,
  linkcolor    = blue,
  citecolor   = red
}
\usepackage{hyperref}
\usepackage{url}
\usepackage{xcolor}
\usepackage{subfig}
\usepackage{amsmath}

\usepackage{color}
\newcommand{\mytilde}{\raise.22ex\hbox{$\scriptstyle\mathtt{\sim}$}}

\newcommand{\changeS}[1]{\textcolor{black}{#1}}

\newcommand{\MS}[1]{\textcolor{blue}{#1}}
\newcommand{\etal}{\textit{et al}.}
\begin{document}

\title{Correct block-design experiments mitigate temporal correlation bias in EEG classification\\ \vspace{0.4cm}
\Large{\textit{A response to the alleged perils and pitfalls of block design~\cite{siskind}}} }
\author{Simone~Palazzo$^*$,
	Concetto~Spampinato$^*$,
	Joseph Schmidt,
	Isaak~Kavasidis, 
	Daniela~Giordano,
	and Mubarak~Shah
	\IEEEcompsocitemizethanks{\IEEEcompsocthanksitem S. Palazzo, C. Spampinato, I. Kavasidis and D. Giordano are with the Department of Electrical, Electronic and Computer Engineering, University of Catania, Viale Andrea Doria, 6, Catania, 95125, Italy.\protect\\
E-mail: {palazzosim, cspampin, kavasidis, dgiordan}@dieei.unict.it
\IEEEcompsocthanksitem M. Shah and C. Spampinato are with the Center of Research in Computer Vision, University of Central Florida. E-mail: shah@crcv.ucf.edu
\IEEEcompsocthanksitem J. Schmidt is  with the Department of Psychology, University of Central Florida,  E-mail: Joseph.Schmidt@ucf.edu
\IEEEcompsocthanksitem $^*$ Equal contribution.
}
}



\IEEEtitleabstractindextext{%
\begin{abstract}
It is argued in~\cite{siskind} that \cite{Spampinato2016deep} was able to classify EEG responses to visual stimuli solely because of the temporal correlation that exists in all EEG data and the use of a block design. While one of the analyses in~\cite{siskind} is correct, i.e., that low-frequency slow EEG activity can inflate classifier performance in block-designed studies~\cite{Spampinato2016deep}, as we already discussed in~\cite{model10}, we here show that the main claim in~\cite{siskind} is drastically overstated and their other analyses are seriously flawed by wrong methodological choices. Our counter-analyses clearly demonstrate that the data in~\cite{Spampinato2016deep} show small temporal correlation and that such a correlation minimally contributes to classification accuracy.
Thus, \cite{siskind}'s analysis \changeS{and criticism} of block-design studies does not generalize to our case or, possibly, to other cases. 
To validate our counter-claims, we evaluate the performance of several state-of-the-art classification methods on the dataset in~\cite{Spampinato2016deep} (after \changeS{properly} filtering the data) reaching about 50\% classification accuracy over 40 classes, lower than in~\cite{Spampinato2016deep}, but still significant. 
We then investigate the influence of EEG temporal correlation on classification accuracy by testing the same models in two additional experimental settings: one that replicates \cite{siskind}'s rapid-design experiment, and another one that examines the data between blocks while subjects are shown a blank screen. In both cases, classification accuracy is at or near chance, in contrast to what~\cite{siskind} reports, indicating a negligible contribution of temporal correlation to classification accuracy. We, instead, are able to replicate the results in~\cite{siskind} only when intentionally contaminating our data by inducing a temporal correlation. This suggests that what Li~\etal~\cite{siskind} demonstrate is \changeS{simply} that \emph{their data are strongly contaminated by temporal correlation and low signal-to-noise ratio}.
We argue that the reason why Li~\etal~in~\cite{siskind} observe such high correlation in EEG data is \changeS{their} unconventional experimental design and settings \changeS{that violate} the basic cognitive neuroscience study design recommendations, first and foremost the one of limiting the experiments' duration, as instead done in~\cite{Spampinato2016deep}.
The reduced stimulus-driven neural activity, the removal of breaks and the prolonged duration of  experiments in~\cite{siskind}, removed the very neural responses that one would hope to classify, leaving only the amplified slow EEG activity consistent with a temporal correlation.
Furthermore, the influence of temporal correlation on classification performance in~\cite{siskind} is exacerbated by their choice to perform per-subject classification rather than the more commonly-used \changeS{and  appropriate} pooled subject classification as in~\cite{Spampinato2016deep}.
Our analyses and reasoning in this paper refute   
the claims of the \textit{``perils and pitfalls of block-design''} in \cite{siskind}. Finally, we conclude the paper by examining a number of other oversimplistic statements, inconsistencies, misinterpretation of machine learning concepts,
speculations and misleading claims in~\cite{siskind}.
\end{abstract}

\def\abstractname{Note}

\begin{abstract}
This paper was prepared as a response to~\cite{siskind} before its publication and we were not given access to the code (although its authors had agreed, through the PAMI EiC, to share it with us). For this reason, in the experiments presented in this work we employed our own implementation of their model.
\end{abstract}

}

\maketitle

\IEEEdisplaynontitleabstractindextext

\IEEEpeerreviewmaketitle

\IEEEraisesectionheading{\section{Introduction}\label{sec:introduction}}
\input{introduction.tex}

\section{Overview}
Here we provide an overview of the data collection effort, the experiments carried out for our counter-analysis, and the main findings:
\begin{itemize}
\item Section~\ref{sec:experiments} describes the EEG data used in this work and the classification performance obtained by state of the art deep models. In particular:
\begin{itemize}
\item Section~\ref{sec:datasets} provides the data collection procedure and pre-processing details of the three datasets used in this work: (a) \textit{Block-Design Visually-Evoked (BDVE) dataset} --- the dataset based on a block design presented in \cite{Spampinato2016deep}; (b) \textit{Block-Design Blanks (BDB) dataset} --- an additional dataset containing EEG recordings of the same subjects who underwent the experiment in \cite{Spampinato2016deep},  while they  were shown a blank screen for 10 seconds between each of the 40 image classes, as per the original acquisition protocol; (c) \textit{Rapid-Design Visually-Evoked (RDVE) dataset} --- EEG data collected during our replication of the rapid-design experiment presented in \cite{siskind}. 

\item Section~\ref{sec:classification} re-analyzes the results of the models proposed in \cite{Spampinato2016deep,decoding_arxiv,model10} as well as of state-of-the-art deep models on  all three  above datasets. The summary of the findings is as follows:
\begin{itemize}
    \item Classification accuracy in the block-design \textit{BDVE} dataset, when the EEG data is correctly filtered, reaches approximately 50\%, which is significant  over 40 classes. This is lower than the $\sim$83\% reported in~\cite{Spampinato2016deep}, but that analysis was biased by slow EEG activity. 
    Thus, we agree that classifying raw EEG data, without any pre-processing, may inflate classifier performance, as stated in~\cite{siskind} and already discussed in~\cite{model10}.
    \item Classification accuracy on the \textit{BDB} dataset is only slightly above chance, suggesting \textit{a negligible temporal correlation in \cite{Spampinato2016deep}'s data.}
    \item Classification accuracy on the \textit{RDVE} dataset when using block-level labels, as done in \cite{siskind}, is also slightly above chance. This stands in stark contrast to the robust classification performance reported in \cite{siskind}. 
    Analogously, performance is at chance when using the correct labels.
    We argue that the high accuracy obtained in \cite{siskind} on the rapid-design experiment with block-level labels is the direct result of a) an incorrect experimental and recording protocol, which increases the experiment duration, removes breaks, and undermines the subjects' attention level, and b) the use per-subject classification.
\end{itemize}
\item In section~\ref{sec:cont}, we attempt to understand why the pattern of classification accuracy that we found is so different from those in \cite{siskind}. In particular, we are only able to reproduce 
\cite{siskind}'s results with our data when we intentionally contaminate it with temporal correlation and magnify its impact on classification results through per-subject analysis as in~\cite{siskind}. This suggests that data in~\cite{Spampinato2016deep} and our new data show a significantly reduced temporal correlation compared to  data in~\cite{siskind}. Again, the reason for the large bias in data in \cite{siskind} is due to an incorrect experimental design that does not follow EEG/ERP design guidelines and to problematic procedures such as per-subject analysis.
\end{itemize}
\item In Section~\ref{sec:disp}, we comment in detail on the experimental design utilized in~\cite{siskind} and the differences, with related consequences on results, from the designs in~\cite{Spampinato2016deep}. We also report what cognitive neuroscience suggests would be a more appropriate design.
\item In section~\ref{sec:incon}, we  carefully analyze~\cite{siskind} and expose the plethora of false claims, misinterpretations of our previous work and of basic machine learning concepts, and unverified opinions which appear to have the sole purpose of hurting the credibility of the authors of~\cite{Spampinato2016deep,brain2image,gan_brain_iccv_2017,tirupattur_acmmm,decoding_arxiv}.
\end{itemize}

\section{Experiments}\label{sec:experiments}
\subsection{EEG data collection and pre-processing}\label{sec:datasets}
\input{dataset.tex}

\subsection{Influence of EEG temporal correlation on  classification tasks}\label{sec:classification}
This section investigates the extent to which temporal correlation in \cite{Spampinato2016deep}'s dataset may have influenced the performance of deep classifiers. 
We first compute EEG classification performance on the block-design BDVE dataset proposed in \cite{Spampinato2016deep} (described in the previous section) to show what is the current state-of-the-art classification performance on this dataset.  Then, in order to  investigate if the obtained results are due to a temporal correlation in EEG data:
\begin{itemize}
    \item We perform new experiments on the \emph{Block-Design Blanks -- BDB -- dataset};
    \item We replicate the experiments on a rapid-design dataset with block-level labels as proposed by \cite{siskind} on \emph{our Rapid-Design Visually-Evoked -- RDVE -- dataset}.
\end{itemize}

We employ the methods presented in~\cite{Spampinato2016deep,model10} and other state-of-the-art EEG classification methods, including \cite{Lawhern_2018} and \cite{NIPS2017_7048}\footnote{Released source code ported by us to PyTorch.}, in order to properly link our previous work to the current state of the art.
Given that Li \etal~\cite{siskind}'s source code is not available, we are unable to replicate the identical analysis steps on our block-design and rapid-design data. 

As mentioned earlier, in our experiments on the block-design dataset we use the dataset splits of \cite{Spampinato2016deep}, with 80\% of the data used for training, 10\% for validation and 10\% for test. 
We employ a standard cross-entropy loss function, minimized  through mini-batch gradient descent using the Adam optimizer, with a learning rate of 0.001, and a mini-batch size of 16. Training was carried out for 200 epochs. Test accuracy is reported at the lowest validation accuracy.

Given that EEG-related object categorization should pool data across subjects~\cite{Stewart2014single}, we compute performance by pooling data from all subjects, instead of per-subject (as was done in Li \etal~\cite{siskind}). Indeed, we will show in Sections~\ref{sec:cont} and \ref{sec:disp} how and why single-subject data tends to maximize the effect of temporal correlation in EEG classification tasks.

\subsubsection{Block-design experiment}
\label{sec:classification_eeg}
\input{eeg_classification.tex}

\subsubsection{Rapid-design experiment}\label{sec:saliency}
\input{rapid.tex}

\subsection{Comparison to~\cite{siskind}'s results}\label{sec:cont}
\input{cont.tex}

\subsection{Discussion}\label{sec:discussion}
Here we summarize the findings presented in this section: 
\begin{itemize}
    \item As already demonstrated in recent work~\cite{model10}, and observed in~\cite{siskind}, the higher performance obtained in~\cite{Spampinato2016deep} was due to incorrectly using EEG raw data, because of the drift present in the DC component of the signal that makes block-design classification easier. 
    \item However, with properly filtered EEG data, classification accuracy reaches about 50\%  on 40 classes, which  is still significant. More importantly, the obtained performance seems to be supported by cognitive neuroscience studies. 
    \item Our counter-analysis, aimed at investigating the presence and influence of temporal correlation on classification performance, reveals that our classification results of the block-design data are unaffected (or only slightly affected). Thus, \emph{the data published in~\cite{Spampinato2016deep} is correct, and importantly, the effect of the temporal correlation claimed in~\cite{siskind} is marginal.}
    \item Our replication of experiments in \cite{siskind}  with rapid-design data with block-level labels shows 
    a small contribution of temporal correlation to classification performance (less than 5\%, as shown in Table~\ref{tab:eeg_black}), that is an order of magnitude less than what \cite{siskind} reports (about 50\%). 
    \item The results presented in \cite{siskind} appear to be a consequence of their failure to follow standard EEG/ERP study design procedures, particularly those related to the  duration of experiments --- recommendations that \cite{Spampinato2016deep} did follow. In addition, they did not replicate the analysis in \cite{Spampinato2016deep}, which pooled data from all subjects. 
    \item We verify that the classification results reported in \cite{siskind} are similar to what we obtain when we intentionally create temporal correlation in our EEG data. This seems to suggest that \cite{siskind}'s data suffers from a strong temporal correlation.
    \item Given that \cite{siskind}'s primary evidence for a temporal correlation in \cite{Spampinato2016deep} stems from the very similar performance of the block-design dataset and the random-design dataset using block-level labels, and that this similarity exists only for their data and not our data, \cite{siskind}'s criticism of the block-design experiment in \cite{Spampinato2016deep} is invalid.
    \item In conclusion, our counter-analyses of the experimental settings and results in~\cite{siskind} suggest a strong temporal correlation in their data due to erroneous experimental procedures and uncovential analysis practices.

\end{itemize}

\section{Cognitive neuroscience experimental designs}\label{sec:disp}
This section elaborates in more depth how and why the experimental design of the studies in~\cite{siskind} are at odds with cognitive neuroscience recommendations and compares them with the designs described in~\cite{Spampinato2016deep} and in this paper.

\input{disp.tex}
\section{Additional inconsistencies and fallacies in \cite{siskind}}\label{sec:incon}
In this section, we comment on a number of inaccuracies and logical fallacies in~\cite{siskind} and rebut the remaining points raised by \cite{siskind} and not covered in previous sections.

\input{inc.tex}

\section{Conclusion}\label{sec:conclusions}
\input{conclusions.tex}

\ifCLASSOPTIONcompsoc
  \section*{Acknowledgments}
\else
  \section*{Acknowledgment}
\fi
The authors would like to thank Dr. Martina Platania for supporting the data acquisition phase, Dr. Demian Faraci for the experimental results, and NVIDIA for the generous donation of two Titan X GPUs.

\ifCLASSOPTIONcaptionsoff
  \newpage
\fi

\end{document}

%% file: introduction.tex
Human neuroimaging research has allowed researchers to non-invasively examine the inner workings of the human mind. Early in cognitive neuroscience research, most works utilized block designs --- the sequential presentation of stimuli of the same type --- because the signal-to-noise ratio is significantly higher than in interleaved rapid-event-related designs~\cite{bandettini2000event,miezin2000characterizing} --- the presentation of stimuli of different types in a random order.
The higher signal-to-noise ratio increases the likelihood that researchers would observe differences between conditions if they existed. Importantly, in cognitive research, it is widely believed that when conditions are presented in blocks, humans tend to react more consistently and respond faster~\cite{ethridge2009consider,roque2016different}, i.e.,  produce more stable responses. 
As a consequence, neuroimaging data should be easiest to classify when a block design is utilized because that is when the strongest signal-to-noise ratio is acquired and when subjects behave the most consistently.
    
A significant amount of neural decoding work has sought to classify human fMRI data~\cite{horikawa2017generic,cichy2016comparison} and this work has generally proven fruitful given the high classification rates achieved by modern deep models. However, fMRI requires large, bulky, costly immobile machines. The availability of lower-cost, mobile systems is a requirement of any Brain-Computer Interface system, hence the recent push to perform a similar style of  classification using electroencephalography (EEG) --- an extensive review of these methods can be found in~\cite{Roy_2019}. However, EEG data tend to be difficult to classify accurately due to the low signal-to-noise ratio. 

In light of the above discussion, an obvious way to improve EEG classification accuracy would be to use a block design to maximize the signal-to-noise ratio of the EEG data. 
This is precisely what~\cite{Spampinato2016deep} and a series of subsequent work~\cite{brain2image,gan_brain_iccv_2017,model10} utilized to build  multi-modal learning methods that combine visual and neural data.
The objective of this line of work was to propose learning methods that would encourage the machine learning community to concentrate more efforts on a new, fascinating, interdisciplinary field; this suggestion was met with a significant amount of interest, spurring several works~\cite{pmid29599461,LI2020107085,NishidaNBMKN20,Kim_2019_ICCV,pmid31354409,8680664}, and, conversely, with skepticism in~\cite{siskind}.  The main criticism in \cite{siskind} is that the use of a block design creates a ``temporal correlation''.  As is well known in EEG research~\cite{luck2014introduction}, temporally nearby data is more similar than temporally distal activity. 
Indeed, temporal correlation is why event-related studies are typically restricted to segments of 1-3 seconds~\cite{luck2014introduction}. Beyond that, slow moving artifactual activity, like skin potentials, creates too much drift in the signal. This slow drift or amplitude fluctuation over time would be responsible for any temporal correlation in the neural data. We assume that it is this electrical amplitude drift overtime that~\cite{siskind} claims is used for classification in block-designed studies. However, temporal correlations seem to exist on the order of a few seconds, rather than minutes (i.e., the case of \cite{siskind,Spampinato2016deep}'s experiments), which would mitigate their impact to classification performance. 

We examine the claims made in~\cite{siskind} and our \changeS{analyses, backed by evidence gathered by a series of experiments,} show that~\cite{siskind} vastly overstated their criticism.
We first verify the EEG classification performance of state-of-the-art deep learning models on the previously-published dataset~\cite{Spampinato2016deep}. Then, we examine the extent to which classification accuracy is affected by a block design, and we show that the temporal correlation in the EEG data~\cite{Spampinato2016deep} leads to a minimal inflation to  classification rates.
 
In the attempt to invalidate the efforts of~\cite{Spampinato2016deep,brain2image,gan_brain_iccv_2017,decoding_arxiv,model10,tirupattur_acmmm}, Li \etal~\cite{siskind} propose a so-called ``rapid-design'' experiment. In this case, stimuli across classes are simply shuffled and presented  in a randomly organized block of trials.  The authors of \cite{siskind} then purport to show that such blocks can be classified based on their temporal order rather than class. 
We argue that this design maximizes the impact of temporal correlations and suppresses neural signals that would be used for classification.
When replicating this rapid design, we found \changeS{that the impact of temporal correlation --- assessed by classifying randomly-ordered stimuli with block-level labels --- is far lower than what was observed by~\cite{siskind}}. Indeed, in our rapid-design study with block-level labels, classification accuracy was inflated by less than  10\%, much lower than the approximately 50\% found in~\cite{siskind}.
We claim that the huge impact of temporal correlation observed in~\cite{siskind} is due to their specific and incorrect experimental design and analysis procedure.
Indeed, Li \etal~\cite{siskind} altered the acquisition protocol used in \cite{Spampinato2016deep}: mainly, they do not report giving subjects any breaks during the experiments. This stands in stark contrast to \cite{Spampinato2016deep}, and also to standard recommendations in cognitive neuroscience experiments~\cite{luck2014introduction}, as the lack of breaks in \cite{siskind} \changeS{entails the strong risk of resulting in inattentive subjects, since the} prolonged exposition to visual stimuli (over 23 minutes, instead of about 4 minutes in~\cite{Spampinato2016deep}) all but guarantees a  vigilance decrement or failure of sustained attention~\cite{jerison1963decrement,nuechterlein1983visual,see1995meta}. 

The consequence of inattentive subjects is that the EEG data do not contain any information related to visual stimuli and the most predominant signal is a slow signal drift resulting in a massive temporal correlation. Indeed, since the data in \cite{Spampinato2016deep} and our new data appear to be minimally affected by temporal correlations, we are only able to replicate~\cite{siskind}'s results  when we intentionally contaminate our data with a bias inducing temporal correlation in EEG signals and perform per-subject classification (as opposed to the pooled settings proposed in~\cite{Spampinato2016deep}). 

In summary, our counter-analysis of \cite{siskind} reveals the following: 
\begin{itemize}
    \item \textit{Technically correct:} As stated in~\cite{siskind} and also in our recent paper~\cite{model10}, slow drift in EEG can inflate the performance of automated classifiers.
    \item \textit{Technically wrong:}  
    The claim in~\cite{siskind} that classification accuracy in block-design experiments is only due to temporal correlation in EEG signals is incorrect. 
    Instead, it appears that the data reported in~\cite{siskind} is highly affected by temporal correlation, due to incorrect experimental protocols, including procedures that maximize the purported temporal correlation and perform per-subject classification.
    However, given the small scale of our and their experiments, the correctness or  incorrectness of what we (and they) found needs to be further examined. 
    \item \textit{Definitely wrong:} We believe that an investigation of the effect of temporal correlations on EEG machine learning classification performance (including possible errors) is part of the natural scientific process that a new --- and immature --- interdisciplinary area must go through.
    However, misinterpreting results, making untrue claims, reporting unverified opinions, and making malicious insinuations\footnote{The pre-print of~\cite{siskind}, available in~\cite{abs-1812-07697}, sports a malicious title (which is repeated several times  in the introduction and conclusion) ---  ``Training on the test set?'' --- with a reference to~\cite{Spampinato2016deep}, which could only cause the reader to question the credibility of the authors of the cited works. We all know how serious an accusation of ``training on the test set'' would be for the machine learning community. Even assuming, \changeS{for the sake of argument,} that there was an inadvertent bias in~\cite{Spampinato2016deep}'s data, that is technically and fundamentally different from `\textit{`training on the test set''}. \changeS{This behavior is reiterated in the abstract of~\cite{siskind}, stating that our block design leads to \textit{``surreptitiously''} training on the test set.}} is unacceptable.
This has nothing to do with scientific progress and makes \cite{siskind}'s efforts appear to be oriented to hurt the reputation of the authors of~\cite{Spampinato2016deep} rather than help to  progress  the field.
\end{itemize}

%% file: dataset.tex
We utilize three  datasets to perform our counter-analyses. We  start with the criticized neural activity dataset published in \cite{Spampinato2016deep}, and then present two new datasets.

\subsubsection{Block-Design Visually-Evoked (BDVE) dataset}

We herein provide an exhaustive description of the dataset in \cite{Spampinato2016deep} and include additional details, in order to make this paper self-contained. 

The recording protocol included 40 object classes with 50 images each, taken from the ImageNet dataset~\cite{imagenet_cvpr09}, resulting in a total of 2,000 images. 
Six participants viewed the visual stimuli in a block-based design. All 50 images corresponding to a single class were shown consecutively in a single sequence. Each image was shown for 0.5 seconds for a total duration of 25 seconds per block/class. After each block/class, a 10-second blank screen (with a black background) was shown, then the next class started. Because of the total time needed for the experiment, its execution was split into \textbf{four sessions} of 10 classes each (about 4 minutes per session). After each session the subjects had time to rest and they continued the experiment whenever they felt ready. Short block durations are critical to keeping participants engaged~\cite{jerison1963decrement,nuechterlein1983visual,see1995meta} and to maximize EEG recording quality \cite{luck2014introduction}. Excessively long block duration is expected to result in extremely low quality data because subjects become inattentive and are likely  to create more movement and ocular artifacts. In turn, this would amplify the contribution of the temporal correlation as that becomes the main signal left in the data.
\emph{It should be noted that \cite{siskind} failed to replicate the block durations from \cite{Spampinato2016deep}: experiments in~\cite{siskind} lasted over 23 minutes, while in~\cite{Spampinato2016deep} block durations were about 4 minutes.}

The BDVE dataset contains a total of 11,964 segments (time intervals recording the response to each image); 36 were excluded from the expected 6$\times$2,000 = 12,000 segments, due to low recording quality or subjects not looking at the screen as determined by the gaze movement data from a Tobii T60 eye-tracker. Each EEG segment contains 128 channels, recorded for 0.5 seconds at 1 kHz sampling rate, represented as a 128$\times$L matrix, with $L\approx 500$ being the number of samples contained in each segment in each channel. The exact duration of each signal may vary slightly, so we discarded the first 20 samples (20 ms) to reduce interference from the previous image and then cut the signal to a common length of 440 samples (to account for signals with $L < 500$). 

For a fair and consistent comparison, when using this dataset for EEG-based classification, we employ the same training, validation and test splits as in ~\cite{Spampinato2016deep}, consisting of 1600 (80\%), 200 (10\%), 200 (10\%) images with associated EEG signals, respectively, ensuring that all signals related to a given image belong to the same split. 

\subsubsection{Block-Design Blanks (BDB) dataset}

As mentioned above, our block-design protocol presented a 10-second blank screen between each class block. For each user, we thus record 360 seconds of brain activity corresponding to the visualization of inter-class blank screens (9 black screens for each of the four sessions), for a total of 6$\times$360 = 1,980 seconds of EEG recording across all six subjects.

To construct the BDB dataset we split such signals into 500-sample long segments with an overlap of 100 samples between consecutive segments. 
From this operation we obtain 864 blank screen samples from all the 10-second segments for each user, for a total of 5,184 segments. 

The data from these blank screens are particularly significant because, as claimed in~\cite{siskind}, any contribution of a temporal correlation to classification accuracy should persist throughout the blank screen interval (i.e., the blank interval should be consistently classified above chance as either the class before or after the blank screen).

\subsubsection{Rapid-Design Visually-Evoked (RDVE) dataset}

We also replicate the experiments performed in \cite{siskind}, where authors  additionally collected data using an interleaved or ``\textit{rapid}'' design, in which images were presented in a random order rather than grouped by class.

For the rapid-design data collection, we employ the same 40 classes as in the BDVE dataset. However, we only utilize 25 images per class to reduce the experiment duration. Images were shown in blocks of 50 images as in~\cite{siskind}, after which a blank screen was shown for 10 seconds. 

Replicating~\cite{siskind}, we did not split the recordings into sessions and performed all of them in a single session that lasted about 11.5 minutes. Note that this is still twice as long as \cite{Spampinato2016deep}'s block-design sessions and less than half as long as \cite{siskind}'s sessions (more than 23 minutes).

The resulting dataset consists of 6,000 EEG signals (1,000 per subject, across six subjects), sharing the same characteristics (sampling rate, temporal length, common-length preprocessing) as the BDVE dataset. Given these similarities, we use the same dataset splits as above.

Furthermore, to assess whether the results of the BDVE dataset (shown in the next section) are related to cognitive/visual information in the EEG signals rather than other factors that could be used by a classifier, (e.g., EEG drift, temporal correlation, current phase, etc.), we also use a variant of this dataset by modifying the class labels as proposed by~\cite{siskind}. Rather than using the actual class label for each stimulus (i.e., image), labels were assigned to each stimulus based on the corresponding temporal order in the sequence (the block order) during the recording protocol. 
This procedure replicates \cite{siskind}'s effort to demonstrate temporal correlation bias in~\cite{Spampinato2016deep}, but our results show a very different pattern of classification accuracy. 
 
\subsubsection{Data preprocessing}

In the following experiments, frequency-filtering is performed by applying a second-order Butterworth bandpass IIR filter (we ran experiments using several cut-off frequencies) and a notch filter at 50 Hz. The filtered signal is then z-scored --- per channel --- to obtain zero-centered values with unitary standard deviations. The maximum low-pass cut-off frequency is 95 Hz, since frequencies above 100 Hz rarely have the power to penetrate the skull.

%% file: eeg_classification.tex
As correctly outlined in Li~\etal~\cite{siskind}, the EEG data filtering procedure was unclear in~\cite{Spampinato2016deep}. We start our analysis on the block-design data by evaluating classification performance when using different EEG frequency ranges. Indeed, the cognitive neuroscience community usually employs five general frequency bands, each associated with a variety of brain functions: delta (0-4 Hz), theta (4-8 Hz), alpha (8-16 Hz), beta (16-32 Hz) and gamma (32 Hz to 95 Hz; gamma band activity can be further split into low- and high-frequency gamma with 55 Hz as cut-off). We discard delta waves as these are typically associated to the sleep phase. 
Thus, we first compute the classification performance at the remaining five frequency bands, and the results are given in Table~\ref{tab:eeg_results_frequencies}.
In all cases, classification performance is lower than in~\cite{Spampinato2016deep}, which reported an average accuracy of about 80\%. As outlined in our recent work~\cite{model10}, the lower accuracy is due to the removal of EEG drift, because the previous work inadvertently utilized unfiltered EEG data.
With correctly filtered data and using high-frequency gamma band activity, \cite{Spampinato2016deep} achieves about 20\% classification accuracy; \cite{Lawhern_2018} and \cite{NIPS2017_7048} reach about 30\% accuracy; our recent approach (i.e., EEG-ChannelNet~\cite{model10}) obtains about 50\% accuracy (chance is 2.5\%). 
Given the timescale of high-frequency gamma, it is unlikely that this result is contaminated by a temporal correlation as the activity takes place on a far shorter timescale than low-frequency activity (i.e., it goes through 55-95 cycles a second, resulting in short-lived temporal correlation).  
The higher visual object classification at high gamma frequencies is consistent with the cognitive neuroscience literature related to attention, working memory and perceptual processes involved in visual tasks \cite{CLAYTON2015188,TALLONBAUDRY1999151,JENSEN2007317}. Based on Table~\ref{tab:eeg_results_frequencies}, in our following analyses we employ the 55-95 Hz frequency range and the wider 5-95 Hz for comparison.

\begin{table*}[]
    \centering
    \begin{tabular}{cccccc}
    \toprule
    {\textbf{Band}} & {\textbf{Frequency - Hz}} & {\cite{Spampinato2016deep}} & EEGNet \cite{Lawhern_2018} & SyncNet \cite{NIPS2017_7048} &  EEG-ChannelNet \cite{model10}\\
    \midrule
    Theta, alpha and beta & 5-32  & 8.6\%  &  13.6\%  & 15.2\%& 19.7\%\\
    Low frequency gamma   & 32-45 & 3.8\%  &   18.3\%     & 10.4\% & 26.1\%\\
    High frequency gamma  & 55-95 & 21.8\% & 31.9\% & 31.8\% & 48.8\%\\
    All gamma             & 32-95 & 13.5\% & 20.8\% & 20.7\%& 40.0\%\\
    All bands       & 5-95  & 11.0\% & 24.9\% & 18.9\% & 31.3\%\\
    \bottomrule
    \end{tabular}
    \caption{EEG classification accuracy using different EEG frequency bands on the BDVE dataset. The  45--55 Hz frequencies are discarded because of the power line frequency at 50 Hz. }
    \label{tab:eeg_results_frequencies}
\end{table*}

Given the above results, it seems highly unlikely that such performance is due to temporal correlation in the EEG data. However, to be certain, we further investigate the extent to which our results may be due to temporal correlation, as argued by Li~\etal~in~\cite{siskind}, rather than to class-level information encoded in the neural signals. 
The neural signals recorded between each pair of classes, i.e.,
the \emph{BDB dataset}, can help address this question. Since the neural data in response to the blank screen is equidistant in time from two classes, a strong temporal correlation would result in significantly greater than chance classification of that data as either the class before or the class after the blank screen. Thus, we verify whether a model trained on the block-design \emph{BDVE} dataset would classify blank-screen segments as either the preceding or subsequent class. Finding near chance level classification accuracy here would indicate little to no impact of a temporal correlation.
To assess the temporal correlation we assign two class labels to each blank segment in the BDB dataset, corresponding to the preceding class and the following class. Then, for each of the models trained on the BDVE dataset and whose results are given in Table~\ref{tab:eeg_results_frequencies}, we compute the classification accuracy of the BDB dataset as the ratio of blank segments classified as either one of the corresponding classes.
Results are shown in Table~\ref{tab:eeg_black}, and reveal that all methods are at or slightly above chance accuracy (i.e., 5\%, since for each segment has two possible correct options out of the 40 classes). \textit{This seems to be a clear indication that temporal correlation in \cite{Spampinato2016deep}'s data is minimal, suggesting that block design experiments (when properly pre-processed) are suitable for classification studies}.

\begin{table*}[]
    \centering
    \begin{tabular}{ccc}
    \toprule
    {Method} & \textbf{All frequencies (5-95)  Hz} & \textbf{High frequency gamma (55-95)  Hz} \\
    \midrule
    \cite{Spampinato2016deep} &7.1\% & 7.9\%\\
	EEGNet \cite{Lawhern_2018} & 3.8\% &4.5\%\\
	SyncNet \cite{NIPS2017_7048} & 9.1\% & 7.5\% \\
	EEG-ChannelNet \cite{model10} & 6.8\% & 8.1\% \\
	\bottomrule
    \end{tabular}
    \caption{Classification accuracy on EEG signals related to blank screen presentation (BDB dataset). EEG data is filtered in the 55-95 Hz band, and in all frequency bands 5-95 Hz. Chance level is 5\%.}
    \label{tab:eeg_black}
\end{table*}

%% file: rapid.tex
The previous analysis reveals that the block-design data in \cite{Spampinato2016deep} is only minimally affected by temporal correlation. To provide more evidence and to compare our results to those reported in \cite{siskind}, we replicate their rapid-design experiment with true and block-level (incorrect) labels. We again employ the methods in~\cite{Spampinato2016deep,model10} and the state-of-the-art approaches, but are unable to include the methods in~\cite{siskind} since code is not available.

Similarly to the block-design study, the rapid data classification accuracy is computed by pooling the data across subjects\footnote{In the next section, we also perform per-subject analyses to compare with \cite{siskind}'s results. We do not perform subject-level analyses here because the focus is on the impact of the temporal correlation in our previous data ~\cite{Spampinato2016deep}.}. We use the same data splits, training procedure and hyperparameters described in Section~\ref{sec:classification}. Again, as above, we report the results on the 55-95 Hz high-frequency gamma band and on the wider 5-95 Hz band. The results are given in Tables~\ref{tab:rapid_class} and~\ref{tab:eeg_rapid_block}, and allow us to draw the following conclusions:
\begin{itemize}
 \item The classification accuracy with correct class labels, is at chance for all tested methods. 
 We argue that this is due to the specific design that tends to inhibit stimuli-related neural responses (see Section~\ref{sec:disp} for a thorough discussion).
 \item The classification accuracy, when using rapid-design data with incorrect block-level labels, is at most 9 percent points above chance, suggesting that the rapid design carries some small temporal correlations.

 \item Importantly, we do not observe the same behavior as reported in Li \etal~\cite{siskind} in which the  rapid-design experiments employing block-level labels showed significantly higher performance: using the 1D CNN method proposed in~\cite{siskind}, they report over 53\% accuracy. Equally interesting, their performance is almost identical to the one reported in~\cite{siskind} for the 1D-CNN method   using their block-design experiments. The fact that we could not replicate these experiments calls~\cite{siskind}'s findings into question.
    
\end{itemize}

\begin{table*}[]
    \centering
    \begin{tabular}{cccccccc}
    \toprule
    {Method} & \textbf{All frequency bands (5-95Hz)} &  \textbf{High gamma frequency (55-95Hz)}\\
    \midrule
    \cite{Spampinato2016deep}       & 3.0\% & 2.5\%\\
	EEGNet \cite{Lawhern_2018}      & 2.7\% & 2.5\%\\
	SyncNet \cite{NIPS2017_7048}    & 2.3\% & 2.1\%\\
	EEG-ChannelNet \cite{model10}   & 2.6\% & 2.8\%\\
	\bottomrule
    \end{tabular}
    \caption{EEG classification accuracy on rapid experiment with \emph{class-level labels} in the wide frequency band 5-95 Hz and in the high gamma frequency band 55-95 Hz (RDVE dataset).}
    \label{tab:rapid_class}
\end{table*}

\begin{table*}[]
    \centering
    \begin{tabular}{cccccccc}
    \toprule
    {Method} & \textbf{All frequency bands (5-95Hz)} &  \textbf{High gamma frequency (55-95Hz)}\\
    \midrule
    \cite{Spampinato2016deep}       & 5.1\% & 5.7\%\\
	EEGNet \cite{Lawhern_2018}      & 6.1\% & 8.2\%\\
	SyncNet \cite{NIPS2017_7048}    & 11.3\% & 11.4\%\\
	EEG-ChannelNet \cite{model10}   & 8.9\% & 8.2\%\\
	\bottomrule
    \end{tabular}
    \caption{EEG classification accuracy on rapid experiment with \emph{block-level labels} in the wide frequency band 5-95 Hz and in the high gamma frequency band 55-95 Hz (RDVE dataset).}
    \label{tab:eeg_rapid_block}
\end{table*}

%% file: cont.tex
In the previous experiments, we verified that the datasets presented in Sect.~\ref{sec:datasets} and classification accuracy of deep models are minimally affected by the temporal correlation claimed by~\cite{siskind}. Then, we note that the results of Li \etal~\cite{siskind} on the rapid-design protocol show a much higher influence of EEG temporal correlation and an almost identical correspondence between their block-design and rapid-design experiments --- correspondence that we do not find in our experiments.
Hence, we now turn to investigate the following two aspects: a) why did Li \etal~\cite{siskind} find such similar classification accuracy between the block-design experiment and the rapid-design experiment with block-level labels? and b) why did Li \etal~\cite{siskind} achieve such high performance on the rapid-design experiment with block-level labels?

We start by comparing the results we obtain on our data to those achieved by, and reported in,~\cite{siskind} on their data. Unlike the previous sections, for this analysis, we also use the 1D-CNN~\cite{siskind} approach\footnote{Our implementation, since \cite{siskind}'s code is unavailable.}. To be consistent with~\cite{siskind}'s analysis, we report the classification accuracy on a per-subject basis\footnote{Please note that averaging the per-subject performance is different from pooling data across all subjects. In the former case, per-subject classification accuracy is calculated and then an average across subjects is computed. In the latter case, data from multiple subjects are pooled together for both training and test (i.e., both are performed on data flattened across subjects).} and compare it with results when the data is pooled across subjects, as per standard protocol~\cite{Stewart2014single}. We also apply the same frequency filtering employed by~\cite{siskind}, i.e., a 14-70 Hz bandpass filter with 50 Hz notch filter.
Training procedure and hyperparameters are the same described in Sect.~\ref{sec:classification}.

Table~\ref{tab:comp_block_rapid} summarizes the comparison between the two experimental settings (i.e., block design and rapid design with block-level labels). 

\begin{table}[]
    \centering
    \begin{tabular}{lcc}
    \toprule
     & Block-design & Rapid-design (block labels)\\
    \midrule
    \multicolumn{3}{c}{\textbf{Classification accuracy on our data}}\\
    \multicolumn{3}{c}{\textbf{Average of per-subject results}}\\
    \midrule
    \cite{Spampinato2016deep}       &  12.2\%    &  5.4\%  \\
    EEGNet~\cite{Lawhern_2018}      &  18.7\%    &  9.1\%  \\
    SyncNet~\cite{NIPS2017_7048}    &  22.4\%    &  10.7\% \\
    EEG-ChannelNet~\cite{model10}   &  22.0\%    &  8.9\%  \\
    1D-CNN~\cite{siskind}           &  33.5\%    &  19.6\% \\
    \midrule
    \multicolumn{3}{c}{\textbf{Classification accuracy on our data}}\\
    \multicolumn{3}{c}{\textbf{Subject pooling}}\\
    \midrule
    \cite{Spampinato2016deep}       &  18.8\%   &  6.9\%\\
    EEGNet~\cite{Lawhern_2018}      &  22.3\%   &  7.4\%\\
    SyncNet~\cite{NIPS2017_7048}    &  22.4\%   &  9.9\%\\ 
    EEG-ChannelNet~\cite{model10}   &  40.5\%   &  15.8\%\\
    1D-CNN~\cite{siskind}           &  26.9\%   &  15.5\%\\
    \midrule
    \multicolumn{3}{c}{\textbf{Classification accuracy as reported in \cite{siskind}}}\\
    \multicolumn{3}{c}{\textbf{ on their own data - Average of per-subject results}}\\
    \midrule
    \cite{Spampinato2016deep}       &  16.0\%   & 17.4\%\\
    1D-CNN~\cite{siskind}           &  53.4\%   & 53.3\%\\
    \bottomrule
    \end{tabular}
    \caption{Classification accuracy comparison between our work and \cite{siskind} on the rapid-design dataset (RDVE) when using block-level labels. EEG data is filtered in the range 14-70 Hz.
    Top part of the table shows the results (as the average over subjects) that we obtain when training the models using single-subject data. The central part shows the results we obtain when pooling multi-subject data. The bottom part shows the performance as reported in~\cite{siskind} on a per-subject analysis (as in the top part of the table).}
    \label{tab:comp_block_rapid}
\end{table}

The comparison leads to the following observations:
\begin{itemize}
    \item When evaluating performance by averaging over per-subject results (Table~\ref{tab:comp_block_rapid}, top), the differences between the classification accuracy on the block-design experiment and on the rapid-design experiment with block-level labels is not as small as in in~\cite{siskind}, where almost no difference is reported between the two settings;
    \item When subjects are pooled (Table~\ref{tab:comp_block_rapid}, middle), such difference becomes higher; e.g., the performance of EEGChannelNet~\cite{model10} --- the model that better learns from data --- rises from 22.0\% (per-subject) to 40.5\% (when pooling data);
    \item The classification performance reported by~\cite{siskind} for the 1D-CNN method (Table~\ref{tab:comp_block_rapid}, bottom) on the rapid-design with block labels is significantly higher than the one we obtain on the same method.
\end{itemize}

It is critically important to note the difference between the results we obtain with pooled data and with per-subject data, especially for the method that yields the best performance, i.e., EEG-ChannelNet~\cite{model10}. Subject pooling shows a robust classification advantage (about 20 percent points difference for the EEG-ChannelNet method~\cite{model10}) on the block design, compared to the rapid-design with block-level labels. 

Lower performance with block-design data on per-subject analysis can be explained as follows:
\begin{itemize}
    \item With limited data from only one subject, the deep models tend to overfit the data more rapidly;
    \item Pooling across subjects may minimize the impact of temporal correlation as the exact timing and intensity of low frequency activity (which is largely responsible for temporal correlations in EEG data) relative to each class would be expected to differ greatly from person to person and session to session. This variance in the training data almost certainly forces the model to learn more general classification rules and to discount the temporal correlation present in any one subject's data.
\end{itemize}

This seems to indicate that the effect of temporal correlation on classification performance is exacerbated by the per-subject analysis proposed in~\cite{siskind}, even though~\cite{Spampinato2016deep} clearly explains that subject data were pooled.
Later in this section and in Section~\ref{sec:disp} we will provide more evidence of the effect of per-subject analysis on classification performance. 
However, while this analysis partially explains why, when pooling data, the difference between block and rapid design results is higher than the one observed in~\cite{siskind}, it does not clarify why~\cite{siskind} showed such strong classification accuracy, significantly higher than what we report here. This is especially true in the rapid-design experiments with block-level labels, as shown in Table~\ref{tab:comp_block_rapid}.

We argue that this is due to~\cite{siskind}'s experimental design. Key differences between Li \etal~\cite{siskind} and~\cite{Spampinato2016deep}'s data collection procedures include the duration of each session, the total duration of the experiment, and the timing and number of breaks provided to subjects. This difference is also mentioned, and evidently underestimated, by~\cite{siskind}:
\begin{itemize}
\item[] \textit{``Unlike the data collection effort of Spampinato et al. [31], which divided each recording into four 350s sessions, each of our 36 recordings was collected in a single session.''}
\end{itemize}

In the data acquisition effort in~\cite{Spampinato2016deep}, subjects were given multiple breaks so that they could remain attentive during the task. If subjects are asked to remain attentive for more than a few minutes, a decrease in vigilance begins and attention starts to wane~\cite{jerison1963decrement,nuechterlein1983visual,see1995meta}. Humans tend to become measurably inattentive in as little as \textbf{8 minutes} and this is especially true during boring passive tasks in which the subject is not asked to respond~\cite{jerison1963decrement,nuechterlein1983visual,see1995meta}, as is the case in \cite{siskind} and \cite{Spampinato2016deep}. This is why the gold standard of EEG/ERP study design suggests that blocks should be between 5-7 minutes long and separated by 1-2 minutes of rest, with additional short breaks of 15-20 seconds, 2-3 times during the block of trials~\cite{luck2014introduction}. The block-designed study in~\cite{Spampinato2016deep} followed such recommendation by using block durations of $\sim$4 minutes, while the block duration in~\cite{siskind}'s experiments was reported to be $\sim$23 minutes ($\sim$28 minutes in the video experiment), with no reported break given to subjects during that time\footnote{It also appears that the temporal correlation was magnified in~\cite{siskind} because they carried out the whole study (not only one single experiment) in a single session. Indeed, the data provided by \cite{siskind}'s authors to us during the PAMI review process,  at \url{http://upplysingaoflun.ecn.purdue.edu/~qobi/tpami2020} seems to show that some subjects underwent the entire series of experiments in a single session; e.g., by looking at file timestamps (the data comes with no metadata providing details on the recording sessions), subject 6 likely underwent experiments for more than \textbf{three hours} (180 minutes) consecutively.}. Given the experiment duration in~\cite{siskind}, it would require near super human attention to stay focused on the task designed in~\cite{siskind}. This would all but guarantee that subjects would ignore the images, become inattentive and daydream, making it impossible to classify the neural data. Whereas this methodological choice would be odd and worrisome in any context, in the current research context it is unclear if this was done by design or not. 
Additionally, the authors of \cite{siskind} do not report using eye gaze data for controlling experiments as we did in~\cite{Spampinato2016deep}. Given this, they have no way to verify if subjects were looking at the images. It would be impossible to classify neural data in response to an image, if the subject had their eye's closed or they were looking away when the stimulus was presented.  Knowing the subjects actually viewed the stimuli is a critical control measure that is missing in~\cite{siskind}.

In the attempt to explain  \cite{siskind}'s results especially on rapid-design with block labels, we carry out an additional experiment by intentionally adding a bias to our EEG data and computing classification accuracy of our implementation of 1D-CNN method~\cite{siskind} and \cite{Spampinato2016deep} (the two deep methods tested in~\cite{siskind}). 
We do this analysis on both the block-design and rapid-design data with block-level labels and using per-subject analysis, as done in \cite{siskind}, and pooled data.

\begin{figure*}
    \centering
    \includegraphics[width=0.33\textwidth]{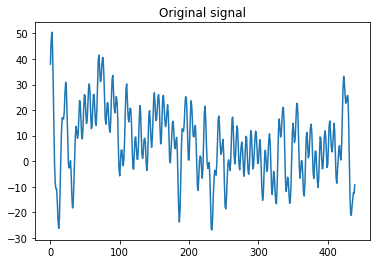}
    \includegraphics[width=0.33\textwidth]{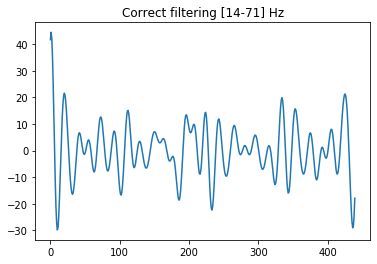}
    \includegraphics[width=0.33\textwidth]{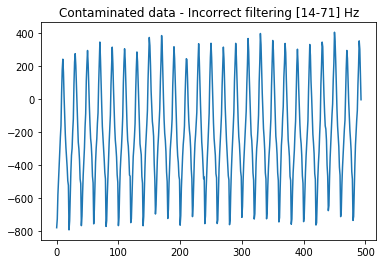}
    \caption{(Left) Original raw EEG signal; (middle) correctly filtered signal in the 14-70 Hz band; (right) incorrectly filtered signal (along the channel dimension) in the 14-70 Hz band, showing temporal correlation.}
    \label{fig:contamination}
\end{figure*}

Specifically, we \emph{contaminate} our dataset by filtering the signals along the channel dimension rather than the temporal dimension (this is \emph{not} a standard procedure in signal processing). The effect of this unconventional filtering is
shown in Figure~\ref{fig:contamination}, where it can be noticed that such contamination replaces the original data with an almost regular pattern over time, emulating a temporal correlation. Incidentally, such contamination can happen by simply transposing the channel and time dimensions in the input EEG data during filtering.

Table~\ref{tab:block_contaminated} shows the performance obtained over the contaminated dataset. Focusing specifically on 1D-CNN performance, the following is observed:
a) its per-subject classification accuracy is sensibly higher than the one with correct data (compare first row in Table~\ref{tab:block_contaminated}  with fifth row of top Table~\ref{tab:comp_block_rapid});
b) its per-subject classification accuracy on the contaminated scenario resembles the one reported in~\cite{siskind} (compare first row in Table~\ref{tab:block_contaminated}  with second row of bottom Table~\ref{tab:comp_block_rapid});
c) in case of huge bias, as the one introduced with the contaminated scenario, pooling data significantly decreases its overall performance (from about \MS{66.8}\% to \MS{8.7}\% on block design and from \MS{47.2}\% to 3\% on the rapid design, shown in the first row of Table~\ref{tab:block_contaminated}).
Moreover, it is important to note that, with the data affected by strong temporal correlation as in our contaminated scenario, pooling data from multiple subjects affects more the accuracy of 1D-CNN method~\cite{siskind} than the one of~\cite{Spampinato2016deep}. We argue this depends on the models' architecture. 
Indeed, 
1D-CNN~\cite{siskind}, 
unlike \cite{Spampinato2016deep}, employs channel-wise 1D convolutions at the initial layers. Allowing the network to focus on each channel independently makes it easier to capture spurious discriminative dynamics, compared to the case when all channels are processed from the first layer of the model. This means that if, for instance, only one channel of one subject shows specific temporal correlation in EEG data, classification results for that subject will be higher and, in case of per-subject analysis, this will affect average classification accuracy.
When pooling data, 1D-CNN~\cite{siskind}, instead, cannot use single-channel specific dynamics and this explains the significant drop in performance reported in Table~\ref{tab:block_contaminated}.
A similar behavior was also observed for SyncNet~\cite{NIPS2017_7048} on the contaminated scenario: its performance drops from 67.4\% with per-subject analysis to 30.3\% with pooled data on block-design and, from 41.2\% to 11.8\% on the rapid design as shown in Table~\ref{tab:remaining}. 
Similar to 1D-CNN~\cite{siskind}, SyncNet~\cite{NIPS2017_7048} performs channel-wise 1D convolutions at the initial layers. The other models~\cite{Lawhern_2018,model10}, instead, show a behavior similar to~\cite{Spampinato2016deep} (i.e., performance with pooled data higher than, or similar to, that with per-subject data, see Table~\ref{tab:remaining}) and are all characterized by having convolutions processing all EEG channels together since the first layer; naturally, 
given the strong induced temporal correlation, those methods are able to achieve good results even when pooling data. 

\begin{table*}[]
    \centering
    \begin{tabular}{cccccc}
    \toprule
    & \multicolumn{2}{c}{\textbf{Block-design}} & & \multicolumn{2}{c}{\textbf{Rapid-design with block labels}}\\
    \cmidrule{2-3} \cmidrule{5-6}
    & Average per subject & Pooling subjects & & Average per subject & Pooling subjects\\
    \cmidrule{2-3} \cmidrule{5-6}
    1D-CNN~\cite{siskind} & 66.8\%& 8.7\%& &47.2\% & 3.0\%\\
    \cite{Spampinato2016deep} & 16.9\%& 21.2\%& &8.8\% & 9.6\%\\ 

    \bottomrule
    \end{tabular}
     \caption{EEG classification accuracy of  \cite{siskind} and \cite{Spampinato2016deep} on the contaminated scenario for block-design and rapid-design with block labels when data is  incorrectly filtered in the 14-70 Hz range.}
    \label{tab:block_contaminated}
\end{table*}

\begin{table*}[]
    \centering
    \begin{tabular}{cccccc}
    \toprule
    & \multicolumn{2}{c}{\textbf{Block-design}} & & \multicolumn{2}{c}{\textbf{Rapid-design with block labels}}\\
    \cmidrule{2-3} \cmidrule{5-6}
    & Average per subject & Pooling subjects & & Average per subject & Pooling subjects\\
    \cmidrule{2-3} \cmidrule{5-6}
    EEGNet~\cite{Lawhern_2018}& 62.6\%& 82.1\%& &68.2\%& 82.6\%\\ 
    SyncNet~\cite{NIPS2017_7048} & 67.4\%& 30.3\% & & 41.2\%& 11.8\%\\
    EEG-ChannelNet~\cite{model10} &80.3\%& 91.6\%& &52.1\%&  71.9\%\\
    \bottomrule
    \end{tabular}
     \caption{EEG classification accuracy of EEGNet~\cite{Lawhern_2018}, SyncNet~\cite{NIPS2017_7048} and EEG-ChannelNet~\cite{model10}  on the contaminated scenario for block design and rapid design with block labels when data is  incorrectly filtered in the 14-70 Hz range.}
    \label{tab:remaining}
\end{table*}

Of course, by including this analysis of ``contaminated data'', we do not intend to make any assumption about the filtering procedure of \cite{siskind}. We simply attempt to demonstrate that when a strong temporal correlation is present, classification accuracy shows the behavior reported in \cite{siskind}, i.e., a smaller difference between block-design results and block-labeled rapid-design results, high classification accuracy in both settings, and per-subject analysis maximizing the impact of temporal correlation on classification results. 
Of course, it would have been ideal to run this analysis on~\cite{siskind}'s data and code, but code was not available and data was provided to us without any additional information to understand the experiments\footnote{\cite{siskind} provided us raw EEG signals in \url{http://upplysingaoflun.ecn.purdue.edu/~qobi/tpami2020}, but it is not even possible to understand what experimental design those data pertain to.}.

%% file: disp.tex
\subsection{Duration of the experiments }\label{sec:comment_duration}

As mentioned in the previous sections, it is unclear why \cite{siskind} performed all experiments in a single session, rather than splitting it into shorter sub-sessions, as is unmistakenly indicated in~\cite{Spampinato2016deep}. The only effect of increasing the sessions duration is to increase fatigue, which, in turn, decreases the subject's vigilance. This is a well-known issue in the cognitive neuroscience and human factors communities ~\cite{jerison1963decrement,nuechterlein1983visual,see1995meta,luck2014introduction}. 
To provide some evidence of this to readers of the machine learning community, we attempt to quantify the impact of experiment duration on temporal correlation in EEG data and how this may inflate classification accuracy. Specifically, we measure the difference between the classification accuracy of 1D-CNN~\cite{siskind} and chance accuracy on the \textit{BDB} (with the same procedure explained in Section~\ref{sec:classification_eeg}) and \textit{RDVE} (with block-level labels) datasets. Indeed, as mentioned earlier, these datasets contain neural responses which are not related to any visual class stimuli. Any classification rate above chance on these data can be accounted for by temporal correlation. 
For this analysis EEG data is filtered between 14-70 Hz and results are computed on a per-subject basis,  consistently with~\cite{siskind}.
We compare our results with those obtained for the same 1D-CNN method in~\cite{siskind}. Specifically, from~\cite{siskind}, we report\footnote{~\cite{siskind}'s accuracy values are taken from Table 9 of \cite{siskind}'s manuscript and from Tables 41--45 of \cite{siskind}'s appendix.} the performance (as average classification accuracy across all subjects)  of the 1D-CNN method on the rapid-design (with block labels, as this is supposed to show temporal correlation) with both images and videos.

Results, in Table~\ref{tab:experiment_duration}, clearly indicate that as the duration of experiment increases, the magnitude of the temporal correlation, and its impact on classification performance, increases substantially, due to subjects becoming inattentive. Additionally, a longer duration experiment would provide more of an opportunity for the signal to drift over time, which when coupled with the per-subject classification exacerbates the issue. \textit{This is a clear indication that the experimental setting in \cite{siskind} maximizes the temporal correlation in their data but likely has minimal impact on a properly-designed study.}

\begin{table}[]
    \centering
    \begin{tabular}{ccc}
    \toprule
    \multicolumn{3}{c}{\textbf{1D-CNN on our data}}\\
    \midrule
    Experiment & Duration (minutes) & Increase over chance\\
    \midrule
    Blank-data (BDB)      &  $\sim$4 mins & 4.4 \\
    Rapid-design (RDVE)     &  $\sim$11 mins & 17.1\\
    \midrule
    \multicolumn{3}{c}{\textbf{1D-CNN performance, taken from ~\cite{siskind}, on their data}}\\
    \midrule
        Experiment & Duration (minutes) & Increase over chance \\
    \midrule
    Image rapid-design &   $\sim$23 mins & 50.3\\
    Video rapid-design &    $\sim$28 mins& 61.9\\
    \midrule
    \end{tabular}
    \caption{\textbf{Temporal correlation w.r.t. experiment duration for the 1D-CNN method~\cite{siskind}}. The longer the experiment lasts, the larger the temporal correlation. The ``Increase over chance'' column reports the increase in percent points of the 1D-CNN method~\cite{siskind} over chance accuracy. 
    Please note that a similar (or lower) level of performance increase of the 1D-CNN method over the chance level is observed for all the other tested methods on our data, see Tables~\ref{tab:eeg_black} and ~\ref{tab:eeg_rapid_block}.}
    \label{tab:experiment_duration}
\end{table}

\subsection{Rapid-design}\label{sec:comment_rapid}

Beside the duration of the experiment and the lack of breaks, the rapid-design experiment, as proposed in \cite{siskind}, seems designed to suppress the very responses that we hope to classify with machine learning methods. Also, the rapid-design experiment  proposed in~\cite{siskind}  creates a whole host of additional issues, since object recognition tends to last many hundreds of milliseconds (especially when the items change rapidly from trial to trial). This means that components such as the P300 and the N400 may  still be processing the item from one class, when an item from the next class is presented~\cite{luck2014introduction}. This signal overlap certainly results in the signal bleeding into the subsequent trial. This is somewhat less of a problem in a block-design study as recognition is known (from decades) to be faster when there is a consistent context \cite{taylor1978identification}. Indeed, the neural processes involved in object categorization are thought to differ greatly between a blocked and interleaved design; when items from a single category are presented in a block, people tend to focus on the similarities of consecutive items, whereas when items from different categories are interleaved (as in \cite{siskind}'s rapid design) people tend to focus on the differences between items \cite{carvalho2017sequence}. 
A fair test of the classification rates of randomly presented classes would utilize designs common to the event-related potentials literature. These designs generally include a jittered (randomly timed) blank screen between each stimulus with some minimum duration. In fact, \cite{siskind} discusses the need to add jitter between items (page 1), but then fails to do so in the study they present. During this jittered blank interval, a long enough duration is typically included to prevent overlap of the neural responses \cite{luck2014introduction}, i.e., activity generally returns to some near baseline level. Additionally, when event-related potentials are examined, the time right before the stimulus is presented is used to perform a baseline correction, which takes the mean activity for a given channel during a pre-stimulus interval and subtracts it from every sample in that channel \cite{luck2014introduction}. This is expected to eliminate any and all temporal correlation effects not due to subject fatigue or inattention (i.e., a drifting electrical signal over time). Importantly, \cite{siskind} failed to include any jitter in their rapid design study. Essentially, all of the features that make a block-design sub-optimal for classification were included to work against classification accuracy in \cite{siskind}'s rapid design study. 
Additionally, when items are presented in a block, it is possible to make the class very salient (i.e., the participant will notice that  they have viewed 50 dogs in a row), whereas the rapid design obscures the point of the study. In this case, if the subjects were even mildly inattentive, they would certainly fail to think about the class of items being presented to them, something that is far harder to miss in the block-design. 
Importantly, obscuring the class in the way that \cite{siskind} did, without requiring an overt response from the subject, calls into question if the subject was even paying attention to the stimuli, whereas an overt response forces the subject to attend to and more fully process the stimuli to the class level \cite{luck2014introduction}. This flawed design is exacerbated by the effects of the aforementioned prolonged exposition to visual stimuli (much longer than in \cite{Spampinato2016deep}, see Table~\ref{tab:experiment_duration}). If the subject is not actively and fully processing the stimulus (i.e., he/she is not paying attention and is, instead, daydreaming), machine learning methods have little hope of being able to classify the stimulus or class from any neuroimaging data.

\subsection{Per-subject analysis}

It is not clear to us why the authors of~\cite{siskind} presented their results separately for each subject, when in~\cite{Spampinato2016deep} the experiments are performed in a multi-subject pooled analysis. 
In Sect.~\ref{sec:cont} we already showed how the per-subject analysis may enhance the effect of temporal correlation on classification results. The per-subject analysis is critical mainly because EEG data are known to be highly replicable within a person \cite{luck2014introduction}, but also highly specific from person to person \cite{luck2014introduction},  \cite{huang2012human}. In fact, the difference in EEG activity is so great from person to person that it has even been proposed as a method for hard-to-mimic and impossible-to-steal biometric identification, and it is able to identify individuals with ~95 to 100 percent accuracy \cite{huang2012human}. This variability can be observed in the results reported in \cite{siskind}. For instance, in the block-design experiments (Table 4 in \cite{siskind}, and Tables 21-25 in \cite{siskind}'s appendix), classification performance of the 1D-CNN method varies from 37.80\% to 70.50\%. Similarly, in the rapid-design experiment with block-level labels (Table 9 in \cite{siskind}, and Tables 41-45 in \cite{siskind}'s appendix), classification accuracy varies from 19.20\% to 84.10\%. Thus, classification performance was not consistent from subject to subject. 
To further elucidate that the per-subject analysis is problematic, we compare the test performance of the 1D-CNN in two settings: 1) the model is trained and tested using single-subject data; 2) the model is trained with data pooled from all subjects and then tested on single subject data. The results of this comparison are given in Table~\ref{tab:variance} and clearly show how pooling data accounts for inter-subject variability by making classifiers less influenced by subject-specific representation. More specifically, Table~\ref{tab:variance} shows that the variability (measured in terms of standard deviation) among per-subject results decreases significantly when a classifier is trained using all subjects' data (on the rapid design with block labels, the standard deviation of performance between subjects drops from 19.0\% in~\cite{siskind} to 3.6\% in our case). Furthermore, this allows the model to focus on inter-subject discriminative features, reducing the bias due to possible temporal correlation in single subject neural responses. 

\begin{table*}[]
    \centering
    \begin{tabular}{cccccccccc}
    \toprule
    & & Subj. 1 & Subj. 2 & Subj. 3 & Subj. 4 & Subj. 5 & Subj. 6 & Average & Std\\
    \midrule
    \multicolumn{9}{c}{\textbf{Results of 1D-CNN as reported in~\cite{siskind} on their data with per-subject analysis}}\\
        Block-design&&  62.7\% & 50.7\% & 50.4\% & 48.1\% & 70.5\% & 37.8\% & 53.4\% & 10.5\%\\
        Rapid-design with block labels& & 50.1\% &	52.2\% & 54.8\% & 19.2\% &84.1\% &59.6\% & 53.3\% & 19.0\%\\
    \midrule
    \multicolumn{9}{c}{\textbf{Our results of 1D-CNN~\cite{siskind} on our data with per-subject analysis}}\\
    Block-design&&  13.1\% & 36.2\% & 42.5\% & 44.4\% & 31.9\% & 33.1\% & 33.5\% & 10.2\%\\
    Rapid-design with block labels& & 21.4\% &	10.7\% & 25.0\% & 30.4\% &20.5\% &9.8\% & 19.6\% & 7.3\%\\
    \midrule
    \multicolumn{9}{c}{\textbf{Our results of 1D-CNN~\cite{siskind} on (our) pooled data from multiple subjects}}\\
    Block-design&&  17.3\% &	25.8\% &	37.0\%	& 37.0\% &	23.9\%	& 20.4\% & 27.9\% & 7.3\%  \\
    Rapid-design with block labels&& 16.7\% & 10.8\% &16.7\% &20.0\% & 18.3\% &10.3\%& 15.5\%& 3.6\% \\
    \bottomrule
    \end{tabular}
    \caption{Variability in classification performance with per-subject analysis and with pooled data.}
    \label{tab:variance}
\end{table*}

Thus, the large inter-subject differences must be overcome for any viable classification method. Importantly, averaged event-related data from a random sample of ~10 subjects tends to look highly similar to another random sample of ~10 subjects \cite{kappenman2011brainwave}, \cite{luck2014introduction}. Failure to pool data across subjects would, again, only serve to increase the impact of temporal correlation. Indeed, it seems that \cite{siskind} did everything possible to maximize the impact of temporal correlation in their results.

\subsection{Discussion}
In conclusion, carrying out cognitive neuroscience experiments is an extremely delicate process and small changes to the design may lead to completely different results and different neural processes being activated. The block-design in~\cite{siskind} differs from the one in~\cite{Spampinato2016deep} in a number of design choices and analysis procedures, including the duration of each group of trials and the number of breaks (20 minute blocks~\cite{siskind} with no mention of a break, compared to the about 4 minutes in~\cite{Spampinato2016deep}). Given the complex, and mostly unknown, nature of the human brain, it is often not trivial to compare results, even with the same exact designs, as outcomes strongly depend on the subjects' conditions. This is especially true with small scale studies as both \cite{Spampinato2016deep} and~\cite{siskind} are. 
Attempting to generalize from findings with different designs, as~\cite{siskind} did, based on the results of completely different studies is definitively wrong, an unconventional practice and a causal fallacy. Indeed, the fact that, in~\cite{siskind}, block-design results are similar to rapid-design (with block labels) results --- similarity on which the whole refutation in~\cite{siskind} is based --- happens only because both their experiments are flawed by the same methodological error (e.g., the duration of the experiments extremely long), and not because of the block-design. 

%% file: inc.tex
\subsection{Incorrect statements in~\cite{siskind} about availability and reproducibility of our code}

In Section~5.5 of \cite{siskind}, the authors claim that they were allegedly \textit{``hindered''} in their attempt to reproduce the results of \cite{tirupattur_acmmm,gan_brain_iccv_2017,decoding_arxiv,brain2image} by the fact, quoting, \textit{``that the authors have declined to release their code to us, despite requests, and the fact that the published papers lack sufficient detail to replicate their models.''}

As for the first two works, the authors of~\cite{siskind} state  \textit{``Palazzo et al. [24] and Tirupattur et al. [34] appear to employ related but somewhat different encoders. We do not comment on these here since we do not have access to this code''} (\textit{``[24]''} and \textit{``[34]''} are reference numbers within \cite{siskind} for \cite{gan_brain_iccv_2017} and \cite{tirupattur_acmmm}). 
Also, in reference to~\cite{tirupattur_acmmm}, Li \etal~\cite{siskind} state that they \textit{``do not have access to these datasets''}. 
This is clearly incorrect. Indeed, one of the co-authors of \cite{siskind}, Hamad Ahmed,  contacted the first author of \cite{tirupattur_acmmm} on December 19, 2018 and asked for the code. \cite{tirupattur_acmmm}'s first author sent him the code and data promptly and made everything available on Github\footnote{\url{https://github.com/ptirupat/ThoughtViz}}. After that, there was no additional communication. 
We believe Li \etal~\cite{siskind} had enough time to run the provided code and analyze the results, instead of speculating on the results obtained in~\cite{tirupattur_acmmm}.  Claiming that they were not given access to the code and the data, is simply incorrect. 
The claim of insufficient detail to reproduce the code and the unavailability of the code described in~\cite{gan_brain_iccv_2017} are, again, not true. Indeed, the architecture of the EEG encoder and the data are the same as in~\cite{Spampinato2016deep}, 
while the code for the conditional GAN (whose architecture is a standard DC-GAN~\cite{RadfordMC15}, as clearly stated in~\cite{gan_brain_iccv_2017} and  for which multiple implementations are available)  is taken from~\cite{Reed2016generative}, which is publicly available\footnote{\url{https://github.com/reedscot/icml2016}}, and as such, couldn't be shared by us. The same discussion holds for~\cite{brain2image} that used state of the art deep models.
This was also communicated to Jeffrey Mark Siskind on December, 19th 2018.
Finally, we did not release the code when the preprint was made available~\cite{decoding_arxiv}, since it was under review at PAMI (and recently accepted~\cite{model10}) and it might have been subject to changes during the peer-reviewing process. However, we believe that enough implementation details were available in the text to reproduce our work, since the dataset is the same of~\cite{Spampinato2016deep}, and therefore  readily available.

The statements by the authors of~\cite{siskind} related to \cite{tirupattur_acmmm,gan_brain_iccv_2017,decoding_arxiv,brain2image} are simply not true and quite bizarre, at the very least, given that a record of communications exists. Both code and data had been made available to them, and they did not even attempt to implement methods when the code could be easily accessed from online sources (which they were informed of) or reproduced through basic programming.

\subsection{Analysis of EEG encoding in~\cite{siskind}} 

The code initially released to replicate the results in~\cite{Spampinato2016deep} did not perfectly match the description of the architecture employed in the manuscript. 
This was due to rewriting the code in Pytorch, since the model's code had been originally written in Torch Lua.

However, the published code was fixed in early 2019 to match the model described in the manuscript --- whose architecture was not debatable despite \cite{siskind} implying (Section~4.6) that we:
\begin{itemize}
\item[] \textit{``describe this (40-way) classifier alternately as a softmax layer, a softmax classification layer, a softmax classifier, and softmax. Colloquial usage varies as to whether or not this implies use of a fully-connected layer prior to the softmax layer.''}. 
\end{itemize}

Any careful reader will notice that Section~3.2 of~\cite{Spampinato2016deep} specifies that \textit{``the encoder network is trained by adding, at its output, a classification module (in all our experiments, it will be a softmax layer)''}; secondly, Table~2 of~\cite{Spampinato2016deep} clearly shows that all output sizes for the encoder configurations are different than 40. Therefore, it is necessary that the ``softmax layer'' must include a layer projecting the dimensionality of the encoder output to the number of classes.

Ignoring these considerations, 
in the same section, Li \etal~\cite{siskind} claim:
\begin{itemize}
\item[]\textit{"we perform all analyses with both the original unmodified code and four variants that cover all possible reasonable interpretations of what was reported in the published papers."}. 
\end{itemize}

\begin{figure*}[h]
\centering
\begin{tabular}{cccc}
\includegraphics[width=0.3\textwidth]{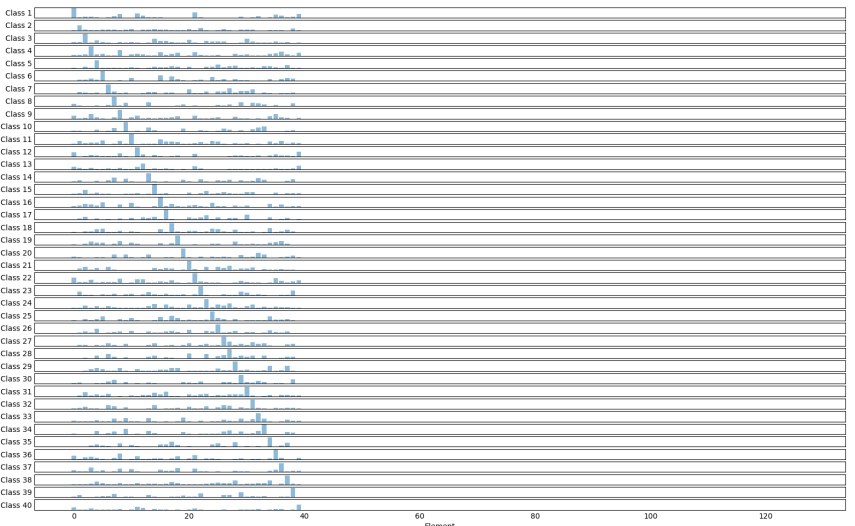} &
\includegraphics[width=0.3\textwidth]{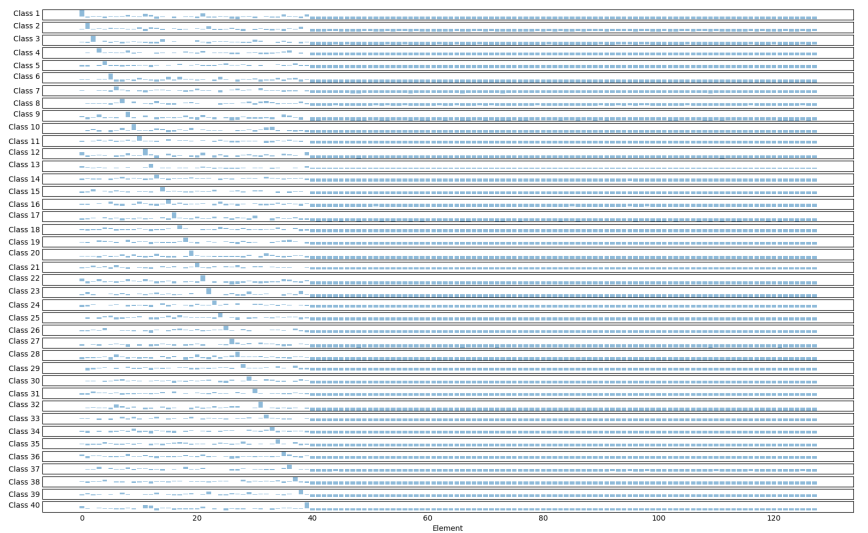} &
\includegraphics[width=0.3\textwidth]{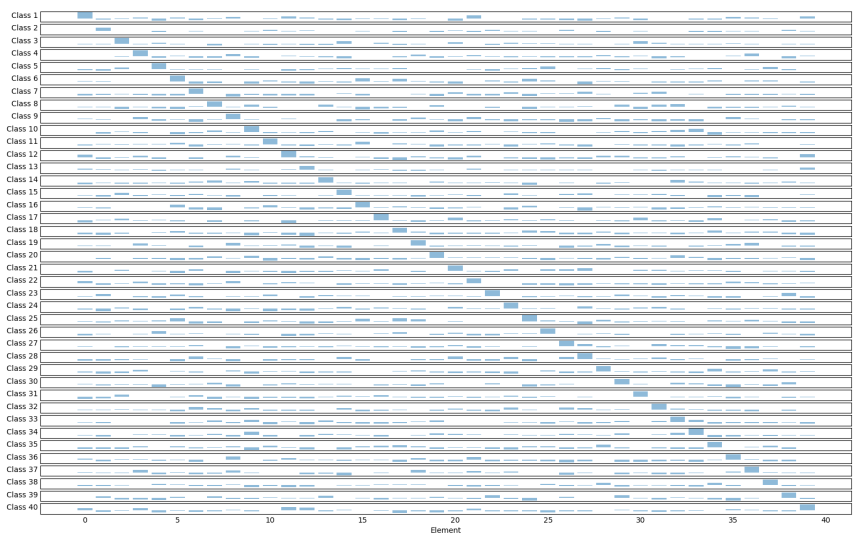} \\
\textbf{(a) LSTM}  & \textbf{(b) LSTM1} & \textbf{(c) LSTM2}  \\[6pt]
\end{tabular}
\begin{tabular}{cccc}
\includegraphics[width=0.3\textwidth]{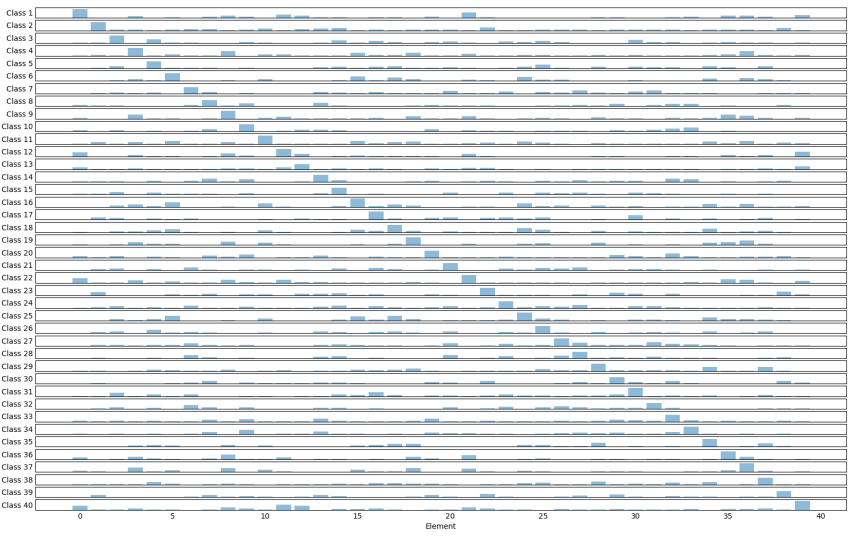} &
\includegraphics[width=0.3\textwidth]{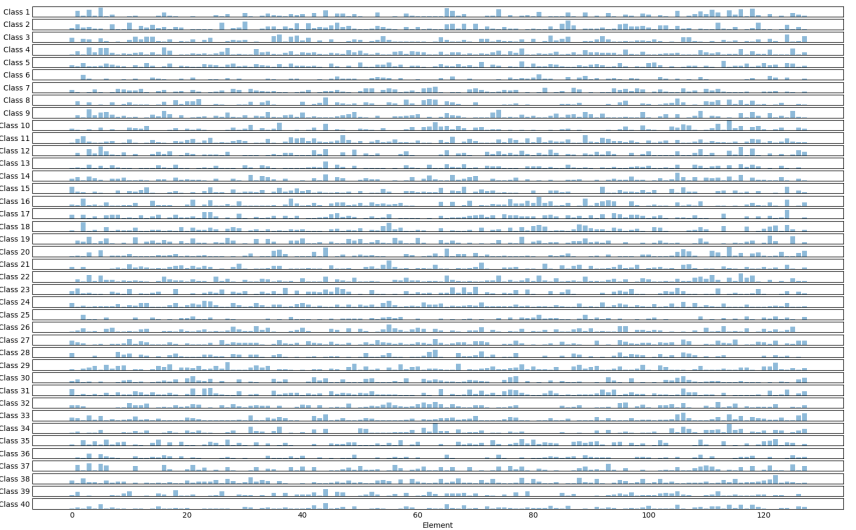} \\
\textbf{(d) LSTM3}  & \textbf{(e) LSTM4}  \\[6pt]
\end{tabular}
\caption{ The average encodings learned by the 5  LSTM variants described in \cite{siskind}. Figures taken from \cite{siskind}'s appendix.}
\label{fig:encodings}
\end{figure*}

Thus, in a fundamentally misplaced effort, Li \etal~in~\cite{siskind} in Sections 1.3 and 1.5 report the learned EEG encodings of five different configurations (that we report here to ease reading):
\begin{itemize}[leftmargin=*]
\item[-] \textbf{LSTM}: LSTM(128) $\rightarrow$ FC(128) $\rightarrow$ ReLU $\rightarrow$ $\|$ $\rightarrow$ cross-entropy. This is the configuration in the code we originally released and then corrected. 
\item[-] \textbf{LSTM1}: LSTM(128) $\rightarrow$ FC(128) $\rightarrow$ $\|$ $\rightarrow$ cross-entropy
\item[-] \textbf{LSTM2}: LSTM(128) $\rightarrow$ FC(40) $\rightarrow$ $\|$ $\rightarrow$ cross-entropy
\item[-] \textbf{LSTM3}: LSTM(128) $\rightarrow$ FC(40) $\rightarrow$ ReLU $\rightarrow$ $\|$ $\rightarrow$ cross-entropy
\item[-] \textbf{LSTM4}: LSTM(128) $\rightarrow$ FC(128) $\rightarrow$ ReLU $\rightarrow$ $\|$ $\rightarrow$ FC(40) $\rightarrow$ cross-entropy. This is the correct configuration as clearly reported in~\cite{Spampinato2016deep}. Indeed, the other three configurations just encode class labels, while this one projects the input EEG data into a new manifold where data can be classified.  
\end{itemize}
The $\|$ indicates the boundary between the encoder and the classifier. Reporting 66 tables for five different configurations, when only one was needed as shown in~\cite{Spampinato2016deep}, appears useless and may only confuse and exhaust the reader without adding anything to the discussion. However, what is irremediably flawed is the line of reasoning in their conclusions, as we will show in the remainder of this section.  Li~\etal~\cite{siskind}, indeed, conclude by stating that all the tested LSTM configurations \textit{``... exhibit the same broad pattern of results''}. 
Figure~\ref{fig:encodings} shows the encodings as computed (and reported) in \cite{siskind}, clearly indicating  that the encodings learned by LSTM4 (which is the exact configuration used in \cite{Spampinato2016deep}) does not show the \textit{``problematic''} one-hot pattern as the other four configurations. 

To justify the different pattern by LSTM4, in  Section~1.5 of the appendix  \cite{siskind}, the authors 
discuss an imaginary issue related to the distribution of the embeddings computed right before the classification layer, complaining that these \textit{``exhibit the same approximate one-hot encoding''} and that it \textit{``means that the encoder output is just a linear mapping of these one-hot encodings''}, i.e., a linear mapping of output classes. 
We believe that the audience of this journal do not need an explanation as to how and why this is wrong, but to prevent anyone from being misled into thinking that such a problem exists, we have to respond. Having a simple linear layer (followed by a softmax activation) after the last hidden layer (with nonlinear activation) of a fully-connected neural network is standard and common practice, e.g., ResNet~\cite{resnet}, Inception/GoogleNet~\cite{inception,googlenet}, DenseNet~\cite{densenet}. 
As already mentioned above, the output layer is only meant to estimate class scores from features that, at that point, are assumed to be linearly separable (hence encode a linear mapping of output classes) if the classifier has learned to classify the input data. 

However, Li \etal~\cite{siskind}, in Section~1.5 of the appendix, attempt to show that the encoder output \textit{``exhibits one-hot pattern"}. In particular, they empirically measure \textit{one-hotness} of feature representation as follows:
\begin{equation}\label{oh}
\text{OH} = |\text{det} (A-I)|, 
\end{equation}
with $A$ being the outer product of the normalized class-averaged encoding vectors and $I$ the identity matrix. Li \etal~\cite{siskind} suggest that all methods for which encoding (pre-classification) features hold the property $\text{OH} \ll 1$, as the method in~\cite{Spampinato2016deep}, are flawed. It is unclear why, according to~\cite{siskind}, having features which can be linearly mapped to one-hot vectors at the model encoding layer should be problematic. 
We argue that the one-hotness behavior described  in~\cite{siskind}, where the outer product of class-averaged features is shown to be close to the identity matrix, is just the expected behavior of any well-trained neural network. 
To demonstrate this, we have prepared an example\footnote{Please find it at \url{https://colab.research.google.com/drive/1Y6HyToZv6HkRKK48D663Fthv8_-4n-nI}}, where a ResNet-18 model is trained on MNIST and the same analysis of pre-output features is carried out. Of course, using ResNet on MNIST is overkill: the choice of the model is meant to show that the phenomenon described by~\cite{siskind} is not an anomaly and is not model-related or embedding-related, while the dataset was chosen to make the example quickly reproducible. Unsurprisingly, the behavior observed with ResNet-18 is the same of the~\cite{Spampinato2016deep}'s EEG encoder as observed by~\cite{siskind}. This is not an issue related to the embeddings, but, again, simply the normal behavior that any machine learning researcher would reasonably expect. 
Beside ResNet-18, we tested several other state of the art models, including InceptionV3, DenseNet-121 and GoogleNet on the MNIST dataset. We measure the one-hotness behaviour claimed by~\cite{siskind} in Table~\ref{tab:encodings_ass}. 
Results show that when a classifier works properly (i.e., it is able to discriminate the input data), $|\text{det} (A-I)| \ll 1$, i.e., features (at the encoding layer) show a \textit{``one-hot representation''}. On the contrary when classifiers have not learned any linear separation among classes (``Untrained models'' in Table~\ref{tab:encodings_ass}), the one-hotness measure is significantly higher than 0, as expected.

\begin{table}[]
    \centering
    \begin{tabular}{ccc}
    \toprule
    Model & \multicolumn{1}{c}{Trained models} &  \multicolumn{1}{c}{Untrained models} \\
    \cmidrule(lr){1-1}
    \cmidrule(lr){2-2} \cmidrule(lr){3-3} 
    & $|\text{det} (A-I)|$ & $|\text{det} (A-I)|$ \\
      \cmidrule(lr){1-1} \cmidrule(lr){2-2} \cmidrule(lr){3-3} 
     ResNet-18  &     3.2 $\times$ $10^{-2}$ & 8.45\\
     DenseNet-121    &2.7$\times$ $10^{-2}$ &	7.52\\
     Inception-v3   & 2.9 $\times$ $10^{-2}$   &  7.53\\
     GoogleNet    &7.8$\times$ $10^{-3}$  &   7.21 \\
     \midrule
    LSTM variants in \cite{siskind}& $|\text{det} (A-I)|$ \\ 
    \cmidrule(lr){1-1} \cmidrule(lr){2-2} 
    LSTM  &   1.2 $\times$ $10^{-4}$ \\
     LSTM1  &   1.9 $\times$ $10^{-1}$ \\
     LSTM2 &   3.1 $\times$ $10^{-4}$ \\
     LSTM3 &   3.7 $\times$ $10^{-3}$ \\
     LSTM4  &  2.4 $\times$ $10^{-4}$ \\
     \bottomrule
    \end{tabular}
    \caption{Comparison, in terms of the one-hotness metric (computed by us) defined in~\cite{siskind}, between state-of-the-art classifiers and the LSTM variants presented in \cite{siskind} (the values in this table are taken from \cite{siskind}'s appendix) . Not trained models column shows the one-hotness metrics for  the same classifiers when no training has been carried out, thus unable to discriminate input image data.}
    \label{tab:encodings_ass}
\end{table}

We carried out the same experiment by adapting the code\footnote{Please find it at \url{https://colab.research.google.com/drive/1aBrz3mbraekDqopFFXIWG1Wm4WiDe36R}. Note that running this code requires  the availability of ImageNet validation data.} to extract features with a pre-trained ResNet-152 model on the ImageNet validation data, achieving $|\text{det} (A-I)| \approx 0$, much lower than those in Table~\ref{tab:encodings_ass}. \textit{Unsurprisingly, state-of-the-art classifiers  exhibit a similar behavior as the LSTM variants described in~\cite{siskind}.
This, according to~\cite{siskind}’s reasoning,
would suggest that state of the art classifiers have not learned anything from image data, which is clearly nonsense.} 

\subsection{\cite{siskind}'s analysis of EEG embeddings}

In Section~3.8 of~\cite{siskind}, the authors argue that since \cite{Spampinato2016deep} regress image features (extracted from a pre-trained CNN) to EEG features (extracted by the encoder trained on EEG signals), this does not imply that these regressed EEG features are meaningful. 

To show this, the authors of \cite{siskind} state that they: 
\begin{itemize}

\item[] \textit{"replace the EEG encodings with a random codebook and achieve equivalent, if not better, results. This demonstrates that the regression and transfer-learning analyses performed by Spampinato et al. [31] are not benefiting from a brain-inspired or brain-derived representation in any way"}. 

\end{itemize}

In particular, Li \etal~in \cite{siskind} generate a synthetic dataset by uniformly sampling 40 codewords --- one for each class --- and generating class samples by adding Gaussian noise to each codeword. They then compute image-specific codewords by averaging across subjects. Note that the resulting dataset is \textit{linearly separable by construction}: the initial 40 codewords represent cluster centers, and the effect of the Gaussian noise is reduced by smoothing across subjects. At this point, they merely show that it is possible to regress, with sufficient accuracy\footnote{We trust the authors' assessment on this ``sufficiency'', since the reported MSE of 0.55, without any information on the size of the codewords and on how/whether the data were normalized, does not say anything about the achieved accuracy.}, a set of class-clustered linearly-separable image features to a corresponding set of class-clustered  linearly-separable codewords, using a linear regressor. This should come as no surprise to anyone. It is even less surprising that replacing the EEG encoder,  that has learned a sub-optimal feature representation w.r.t. the image encoder,  with a the codebook that, instead, has been generated to have an optimal class representation, yields enhanced classification accuracy.

The key observation is that performing this kind of regression is possible \textit{only if the target features encode information that allows them to be clustered by class}, which is obviously true in the codebook experiment described in~\cite{siskind}, since it is \textit{designed} that way, but which is not obvious when those target features are learned by another model (in our case, the EEG encoder). Since it seems possible to learn meaningful (and linearly separable) representations from neural signals, as shown in Table~\ref{tab:eeg_results_frequencies}, there can be no doubt that the regressed features are brain-inspired, simply because the data source they are extracted from is brain activity signals.

The invalid reasoning in~\cite{siskind} can also be shown with an example. Let us suppose that we replace the EEG encoder, in the regression schema in~\cite{Spampinato2016deep}, with a ResNet. Then, we regress visual features extracted from GoogleNet to those learned by a ResNet. 
Of course, within the limits of intra-class variability, such a regression succeeds\footnote{We implemented this experiment and it can be found here: \url{https://colab.research.google.com/drive/1et3Pnlv9Iivtlcku8ck9-KEe2cKHijmv}. To evaluate the regressor, we compared the classification accuracy on a random subset of the ImageNet validation set when using the original features and when using the regressed features. Those two values were respectively about 78\% and 65\%, showing a performance drop that can be expected, given the approximation introduced by regression and the original difference in classification performance between GoogleNet and ResNet (about 9 percent points).}, since a class mapping exists between the two groups of features. 
Then, if we replace ResNet with the codebook proposed in \cite{siskind}, the regression would still work, as implied by the fact that the authors can successfully carry out that regression from VGG-16, that has similar performance as GoogleNet\footnote{See \url{https://pytorch.org/docs/stable/torchvision/models.html}.}. This, according to the reasoning in~\cite{siskind}, should imply that ResNet does not derive any image-related features, which is of course absurd.
What the authors of~\cite{siskind} overlook is the fact that, if ResNet had not learned image-related features in the first place, then the resulting representations would not be class-separable and the regression would fail. Similarly, if the EEG encoder in~\cite{Spampinato2016deep} had not learned EEG-related features, the regression would also simply fail.

Thus, the reasoning done in~\cite{siskind} on the EEG embeddings in~\cite{Spampinato2016deep} looks very much like a straw man argument: indeed, contrary to what is stated in \cite{siskind}, we did not say in~\cite{Spampinato2016deep}  that the regression from pre-trained visual features to the EEG features should result in a benefit to classification accuracy. On the contrary, we clearly show in \cite{Spampinato2016deep}  (in Table 5)  that performance drops when moving from visual to brain-derived features. What is clear from \cite{Spampinato2016deep} is that the regression from visual features to brain-derived features is a simple but effective strategy to employ EEG features that are useful for visual categorization,  and not to improve performance of visual classifiers. This point cannot be refuted. 

\subsection{Comment on \cite{siskind}'s analysis of image generation}

In Section~5.3 of~\cite{siskind}, the authors argue that the generative model in~\cite{tirupattur_acmmm} shows signs of model collapse and it is very unlikely that the images in the paper are generated by a GAN model.
These are unverified claims that seem to have a malicious intent rather than offer scientific findings. Unlike what is assumed by these unsubstantiated claims, in \cite{tirupattur_acmmm}: a)  a widely-employed metric such as the Inception score is used to compare a trained model with existing GAN models, and b)  the correct behavior of the model is verified by interpolating the input EEG encondings between two classes of images, which is a standard way to check for GAN mode collapse~\cite{goodfellow2014generative,radford2015unsupervised}.
Additionally, source code and evaluation routines have been publicly available since December 2018. Li \etal~\cite{siskind} had been informed of the code availability but opted to make their claims without bringing any evidence to substantiate them. 

\subsection{\cite{siskind}'s interpretation of brain-visual joint embedding}

In Section~5.4, the authors of~\cite{siskind} discuss the method and results presented in a preprint~\cite{decoding_arxiv}, now published in~\cite{model10}. 
We would like to point out some corrections to their arguments.

Authors in \cite{siskind} claim the following: 
\begin{itemize}
\item[] \textit{``The loss function employed in the joint training regimen simply constrains the two encoded representations to be similar. A perfectly trained image encoder, trained against class labels, would simply encode image class, no more and no less. A perfectly trained EEG encoder, trained against class labels, would simply encode stimulus class, no more and no less. During joint training of the EEG encoder, the image encoder serves simply as a surrogate for class labels, no more and no less. Similarly during joint training of the image encoder, the EEG encoder serves simply as a surrogate for class labels, no more and no less. Thus joint training accomplishes nothing that could not be accomplished by training the components individually against class labels. The resulting encoded representations would contain no information beyond class labels''}.
\end{itemize}

However, in \cite{decoding_arxiv,model10} we clearly propose a model and a training strategy aimed at: a) preventing the learned features from being \textit{``strongly tied to the proxy task employed to compute the initial representation (e.g., classification)''}, and b) focusing more on learning \textit{``class-discriminative features than on finding relations between EEG and visual patterns''}.
In particular, these papers describe a deep learning architecture to maximize the similarity between brain and visual data without any class supervision, i.e., exactly the opposite of what was claimed in~\cite{siskind} (again straw man argumentation).
The implementation of the method trains the EEG encoder from scratch, hence its initial output will be basically random and will definitely \textit{not} encode a class distribution.  From training practical purposes, \cite{decoding_arxiv,model10} use a pre-trained image encoder and the image classifier's complexity is generally higher than the complexity of the EEG encoder. 
However, even if the image encoder is pre-trained for classification, the encoders resulting from joint training will not necessarily provide class embeddings. On the contrary, since the training loss is based on feature similarity (and not on image classes), and since the EEG encoder will initially provide random values, the image features shift from being class-discriminative to whatever embedding the EEG and image encoder converge to.

\textit{Thus the claim made by Li et al.~\cite{siskind} holds if, and only if, the two encoders are trained against class labels, but as \cite{decoding_arxiv,model10} clearly state, the EEG encoder is trained from scratch and not against class labels, thus invalidating their assumption.}

\subsection{Comment on \cite{siskind}'s analysis of the state of the art}

In Section~6 of \cite{siskind}, the authors discuss the state of the art of \textit{``the field''} (it is not clear to what they are referring to but it sounds as if it applies to all EEG signal analysis in general).

To summarize, they scrutinized 306 papers cited by Roy~\etal~\cite{roy2019deep} or by  \cite{brain2image,gan_brain_iccv_2017,Spampinato2016deep,tirupattur_acmmm}. Excluding self-citations and works that do not collect or use EEG data for classification, a total of 122 papers remained, for which the authors \cite{siskind} state (quoting):
\begin{itemize}
\item[] \textit{``... attempted to assess whether samples in the test set were recorded in close proximity to samples in the training set or the classes in the test trials appeared in the same order as classes in the training trials''}.
\end{itemize}
They come to the following conclusions:
\begin{itemize}
\item[]\textit{``About a third of the papers scrutinized appeared to be clear of the concerns raised here. About another third appeared to suffer from the concerns raised here. It was not possible to assess the remaining third''}.
\end{itemize}

We cannot refrain from pointing out the qualitative nature of this kind of analysis of the state of the art. No individual paper in this analysis is cited, either in the manuscript or in the appendix. No examples of papers belonging to the three ``about-a-third`` categories are cited, except the ones of which we are authors or coauthors. Attempting to assess the veracity of those claims requires one to search the relevant papers and repeat every single study.

In our experience, this is the first time we have seen a journal paper with a ``related work'' section that does not actually cite any papers and expects the readers to either trust the authors or research the subject by themselves. This is also the first time that we have seen a paper discussing a problem that apparently is general (\textbf{``about a third''} of 122 papers suffers from it) by focusing on just a few papers that share a subset of common authors. What appears even more puzzling is that while \cite{gan_brain_iccv_2017,Spampinato2016deep,brain2image,decoding_arxiv} share the dataset being criticized in~\cite{siskind}, the dataset employed in~\cite{tirupattur_acmmm} is from~\cite{Kumar}, which is a completely different dataset, and for which Li \etal~\cite{siskind} clearly claim not to have information\footnote{They clearly state in relation to \cite{Kumar}'s dataset: \textit{"We have no way of knowing whether the test sets contained segments from the same blocks that had segments in the corresponding training sets."}}.
The only thing these papers have in common, is, again, a subset of authors. 

\subsection{Comment on code and data availability of \cite{siskind}}

In our experiments, we tried to replicate the results presented in~\cite{siskind} by evaluating our implementation of their model on our data. Unfortunately, at the time of writing, ~\cite{siskind}'s code was not provided to us and no information about training procedures (hyperparameters, optimizer, etc.) is available in the manuscript.  Given this, we implemented the model to the best of our knowledge based on what could be assessed from their paper and by making reasonable, unbiased, assumptions for missing information. 
Analogously, Li \etal~\cite{siskind} provided us with only raw EEG data but without any information to understand from what experiment each data comes from. Indeed, the raw EEG data contains only a suffix from 1 to 8 (and this does not hold for all subjects) which is intended to describe the specific experiment. 
Moreover, even assuming we could run all the combinations to understand their data and make a guess based on the obtained results, for the rapid experiments the sequence of visual stimuli is unknown.
Lack of code and data hindered us to run experiments and check thoroughly the results reported in~\cite{siskind}. 
For the sake of fairness, we believe this is a point for the readers to be aware of.

%% file: conclusions.tex
In this paper, we have carried out our analyses to respond to the criticisms raised by~\cite{siskind} and these are the main findings:

\begin{itemize}
    \item The claim on the bias introduced by block-design in~\cite{siskind} cannot be generalized to our case, while it appears that the data in \cite{siskind} is highly contaminated by temporal correlation;
    \item The reason for such a huge bias in~\cite{siskind} is due to a failure in replicating the block design in \cite{Spampinato2016deep} (which instead is compliant with  EEG/ERP  study design recommendations). Drastically increasing the session duration (over 23 minutes compared to the about 4 minutes in~\cite{Spampinato2016deep}) can only reduce the subjects' vigilance level, increasing the presence of the temporal correlation in the EEG data and its effect on classification performance;
    \item The rapid-design experiment and settings proposed in \cite{siskind}, are rather uncommon in cognitive neuroscience and are bound to maximize the temporal correlation; 
    \item The temporal correlation bias on classification results is further exacerbated by the per-subject analysis in~\cite{siskind}, in contrast to~\cite{Spampinato2016deep}; 
    \item Thus, \cite{siskind} did fully not replicate~\cite{Spampinato2016deep}'s settings, both from cognitive neuroscience and from machine learning perspective, by  a) altering the experiments and session duration and performing EEG analyses per-subject rather than pooled data analysis, and b) incorporating a new experimental design (i.e., the rapid-design) that does not have any scientific foundation and differs significantly, in terms of expected neural responses, from~\cite{Spampinato2016deep}. 
    \item The analysis by \cite{siskind} of the works~\cite{gan_brain_iccv_2017,brain2image,tirupattur_acmmm,decoding_arxiv} is flawed due to  wrong and unsubstantiated claims, misinterpretation of results and of basic machine learning concepts. 
    \item These above two issues, combined to the rather qualitative analysis of the state of the art, focusing only on a few works sharing the same subset of authors (us), and to the  speculative nature of multiple arguments, calls into question the actual intention of \cite{siskind}.

\end{itemize} \vspace{0.4cm}

The authors of \cite{siskind} start their conclusion with this statement (quoting): 
\begin{itemize}
\item[] \textit{``...the enterprise of using neuroimaging data to train better computer-vision systems [...] requires more sophisticated
methods than simply attaching a regressor to a pretrained object classifier and is also likely to be difficult and beyond the current state of the art.''}.
\end{itemize}

First, for the sake of fairness and correctness, we would like to point out that the approach in~\cite{Spampinato2016deep} is not \textit{``simply attaching a regressor to a pretrained object classifier''}. Such a definition deliberately ignores the overall architecture of EEG feature learning.
Second, it is well-known in the machine learning community that the success of the classification depends mainly on the quality of the representation. Thus, even \textit{``very simple regressors''} may work if the features are descriptive enough of the input data. 
Third, \cite{siskind} includes again unverified opinions that are simply not consistent with the state of the art. For example, Li \etal~\cite{siskind} ignore two recent works by Kamitani et al.~\cite{horikawa2017generic} (published in \textit{Nature Communication}) and Cichy et al.~\cite{cichy2016comparison} (appeared in \textit{Nature Scientific reports}) that employ linear regression models to learn cross-modal neural and visual information for classification. In particular,~\cite{horikawa2017generic} used linear regression from fMRI data to visualize representations learned by a pre-trained object classifier --- thus contradicting \cite{siskind}'s very claim.\\
The conclusion of \cite{siskind} then proceeds with these statements (quoting):
\begin{itemize}
\item[] \textit{"When widely published [11, 18, 20, 24, 25, 30, 31, 34], inordinately optimistic
claims can lead to misallocation of valuable resources and can sideline more modest but legitimate and important advances in the field. Thus, when the sensational claims are recognized as false, it is imperative that the refutation be widely publicized to appropriately caution the community. Further, when the community as a whole appears to suffer from widespread but problematic practices, it is even
more imperative that this warning be widely publicized to appropriately caution the community."}
\end{itemize}
We tend to agree with these statements provided due clarifications:
\begin{itemize}
    \item We agree that inordinately optimistic claims may led to misallocation of valuable resources; analogously, we believe that refuting claims should be grounded to solid scientific practices and it cannot be based on small-scale experiments (as theirs and our work was). Indeed, in the experimental sciences, it is well known that to demonstrate a null result, one needs a considerable amount of data, i.e., the burden of proof to show that something is absent is higher than the burden of proof to show that something is present, because it is always possible that a larger sample size would reveal differences~\cite{cohen1992statistical},~\cite{dienes2014using}. 
    \item The above point highlights the importance of moving the discussion of the cognitive experiment design to a more appropriate community, able to evaluate properly and more critically design choices. Attempting to substantiate the validity of cognitive designs (which we have demonstrated being wrong) with hundreds of tables to give the impression of soundness, as~\cite{siskind} did, is definitively incorrect. Indeed, cognitive neuroscientists would immediately spot the differences between the experimental designs in~\cite{siskind} and~\cite{Spampinato2016deep}, and the consequences on study outcomes of such differences.   
    \item The authors of \cite{siskind} mention \textit{``more modest but legitimate and important advances in the field''}, but we did not see any of these advances discussed in~\cite{siskind}.
    \item The authors of \cite{siskind} also claim \textit{``when the community as a whole appears to suffer from widespread but problematic practices''}, but again in \cite{siskind} we did not see any analysis of \textit{``community practices"} (see Related Work section in \cite{siskind})  beyond ~\cite{Spampinato2016deep,gan_brain_iccv_2017,brain2image,tirupattur_acmmm,decoding_arxiv}. Generalizing to the whole community based on the analysis of a few works and on small scale experiments, designed in opposition to standard cognitive neuroscience practices, is definitively wrong.
    \item We believe that these works~\cite{Spampinato2016deep,gan_brain_iccv_2017,brain2image,tirupattur_acmmm,decoding_arxiv} should be taken as they are, i.e., multimodal learning methods aimed at integrating 2D data with 1D noisy EEG time series. We are aware of the limitations of these experiments from a cognitive perspective. Decoding the human brain surely requires large interdisciplinary efforts. What we can do as machine learning researchers is to propose models and release code and data. And this is exactly what we have done in~\cite{Spampinato2016deep}.
    \item We recognize the importance of \textit{``appropriately caution the community''} (it unclear what community they are referring to: PAMI or cognitive neuroscience?) to prevent any possible errors or biases, but at the same time \textit{"the community"} must be aware of the counter-effects of papers like \cite{siskind} that appear to simply attack the credibility of researchers using mostly speculative and barely scientific claims. This will only serve to discourage researchers from releasing code and data or even from working in this new AI/cognitive neuroscience field.
    A while ago, \cite{CVPR.2011.5995347} showed that the Caltech-101 dataset~\cite{1384978} is extremely biased. If a paper like~\cite{siskind} had been published, criticizing the Caltech-101 dataset, the machine learning community would have lost out on all the valuable classification methods that came from the Caltech-101 dataset. Instead, these methods have contributed significantly to the advancement of image classification research, despite the quality of the dataset.

\end{itemize}

In conclusion, in this paper, we have responded to the main scientific question of the critique, which is unfortunately a very small portion of the largely speculative comments and fallacious argumentation made by~\cite{siskind}.  However, we believe that the additional analysis we carried out has showed that, 
when done properly, block designs are totally appropriate to use and this is largely known in the cognitive neuroscience community.  Furthermore, our analysis has highlighted two main lessons  for mitigating the risks of a temporal correlation bias on EEG classification tasks in block-design experiments:
\begin{enumerate}
    \item To limit the duration of recording sessions to less than 8 minutes, as suggested by cognitive neuroscience literature~\cite{jerison1963decrement,nuechterlein1983visual,see1995meta} and done in~\cite{Spampinato2016deep}, as opposed to the over 20 minutes in~\cite{siskind}, in order to avoid the  subjects becoming inattentive and their related neural responses not containing visual stimuli content;
    \item To pool data from multiple subjects, as done in~\cite{Spampinato2016deep} and opposed to the per-subject analysis proposed in~\cite{siskind}, in order to avoid models to learn spurious specific-subject dynamics rather than a shared representation pertaining the shown stimuli.  
\end{enumerate}

Furthermore, the data in~\cite{Spampinato2016deep} appears to be clean from the temporal correlation claimed in~\cite{siskind} and, as such, it is still a valuable resource for developing and testing multimodal learning methods. 

Of course, to effectively advance the field on decoding human brain, a deeper joint effort between machine learning and cognitive neuroscience communities is required as wrong designs of cognitive experiments may lead to erroneous outcomes (as we have shown), regardless of the quality of machine learning methods.